\documentclass{article}

\pdfoutput=1

\usepackage[nonatbib, final]{neurips_2023}
\usepackage[square,numbers]{natbib} 
\usepackage[utf8]{inputenc} 
\usepackage[T1]{fontenc}    
\usepackage{hyperref}       
\usepackage{url}            
\usepackage{booktabs}       
\usepackage{amsfonts}       
\usepackage{nicefrac}       
\usepackage{microtype}      
\usepackage{xcolor}         

\usepackage{algorithm}
\usepackage{algpseudocode}

\title{Weakly Supervised Detection of Hallucinations \\ in LLM Activations}

\author{%
  Miriam Rateike \\
 Saarland University\\
 IBM Research Africa\\ 
 Nairobi, Kenya
  \And
  Celia Cintas\\
  IBM Research Africa \\ 
  Nairobi, Kenya\\
  \AND
  John Wamburu \\
  IBM Research Africa \\ 
  Nairobi, Kenya \\
  \And
  Tanya Akumu \\
  IBM Research Africa \\
  Nairobi, Kenya  \\
  \And
  Skyler Speakman \\
  IBM Research Africa \\ 
  Nairobi, Kenya \\
}

\newcommand{\Bcal}{\mathcal{B}}

\newcommand{\Tcal}{\mathcal{T}}

\usepackage{amsmath}
\usepackage{graphicx}

\usepackage{multirow} 
\usepackage{caption}
\usepackage{subcaption} 
\usepackage{algorithm} 
\usepackage{wrapfig}
\usepackage{amsthm}
\usepackage{amsmath}
\usepackage{amsfonts}

\usepackage{tikz} 
\usetikzlibrary{bayesnet}
\usetikzlibrary{arrows}

\usepackage{caption}
\usepackage{subcaption}

\usepackage{tikz}
\usetikzlibrary{bayesnet}
\usetikzlibrary{arrows}

\usepackage{tikz}

\usetikzlibrary{shapes,decorations,arrows,calc,arrows.meta,fit,positioning}
\tikzset{
    -Latex, auto, node distance = 0.5 cm and 0.5 cm, semithick,
    state/.style = {circle, draw, minimum width = 0.7 cm},
    const/.style = {minimum width = 0.7 cm},
    inter/.style = {rectangle, draw, minimum width = 0.7 cm, minimum height = 0.7 cm},
    point/.style = {circle, draw, inner sep = 0.04cm, fill, node contents = {}},
    bidirected/.style = {Latex-Latex,dashed},
    el/.style = {inner sep=2pt, align=left, sloped}
}

\usepackage{wrapfig}
\usepackage{bbm}

\usepackage{booktabs}

\usepackage[utf8]{inputenc} 
\usepackage[T1]{fontenc}    
\usepackage{hyperref}       
\usepackage{url}            
\usepackage{booktabs}       
\usepackage{amsfonts}       
\usepackage{nicefrac}       
\usepackage{microtype}      
\usepackage{xcolor}         

\usepackage{amsmath} 
\usepackage{enumitem} 

\usepackage{wrapfig}

\usepackage{caption}

\usepackage{algorithm}
\usepackage{algpseudocode}

\everypar{\looseness=-1}
\usepackage[utf8]{inputenc} 
\usepackage[T1]{fontenc}    
\usepackage{hyperref}       
\usepackage{url}            
\usepackage{booktabs}       
\usepackage{amsfonts}       
\usepackage{nicefrac}       
\usepackage{microtype}      
\usepackage{xcolor}         
\usepackage{amsthm}         

\usepackage{mathtools}
\usepackage{xspace}
\usepackage{amsmath}
\usepackage{bbm}
\usepackage{amssymb}
\usepackage{pifont}
\usepackage{caption}
\usepackage{subcaption}
\usepackage{graphicx}
\usepackage{wrapfig}
\usepackage{enumitem}
\usepackage{multirow}
\usepackage{comment}

\newcommand{\cls}{{\texttt{[CLS]}} }
\newcommand{\scanlr}{{\texttt{scanLR}}}
\newcommand{\scanl}{{\texttt{scanL}} }
\newcommand{\scanr}{{\texttt{scanR}} }

\newcommand{\scanlrshort}{{\texttt{LR}}}
\newcommand{\scanlshort}{{\texttt{L}} }
\newcommand{\scanrshort}{{\texttt{R}} }

\newcommand{\scantwo}{{\texttt{scan2}}}

\newcommand{\clfshift}{{\texttt{clf+}}}
\newcommand{\clfstar}{{\texttt{clf}$\star$ }}

\newcommand{\hallucinations}{{Hallucinations}}
\newcommand{\stereoset}{{Stereoset}}
\newcommand{\realtoxicity}{{RealToxicityPrompts}}

\usepackage{siunitx}
\usepackage{booktabs}
\usepackage{etoolbox}

\renewrobustcmd{\bfseries}{\fontseries{b}\selectfont}
\renewrobustcmd{\boldmath}{}

\newrobustcmd{\B}{\bfseries}

\newrobustcmd{\BB}{\bfseries}

\begin{document}

\maketitle
\begin{abstract}
We propose an auditing method to identify whether a large language model (LLM) encodes patterns such as hallucinations in its internal states, which may propagate to downstream tasks. 
We introduce a weakly supervised auditing technique using a subset scanning approach to detect anomalous patterns in LLM activations from pre-trained models.
Importantly, our method does not need knowledge of the type of patterns \emph{a-priori}.
Instead, it relies on a reference dataset devoid of anomalies during testing. 
Further, our approach enables the identification of pivotal nodes responsible for encoding these patterns, which may offer crucial insights for fine-tuning specific sub-networks for bias mitigation.
We introduce two new scanning methods to handle LLM activations for anomalous sentences that may deviate from the expected distribution in either direction.
Our results confirm prior findings of BERT's limited internal capacity for encoding hallucinations, while OPT appears capable of encoding hallucination information internally.
Importantly, our scanning approach, without prior exposure to false statements, performs comparably to a fully supervised out-of-distribution classifier. 
\end{abstract}

\section{Introduction}
The rapid proliferation of Large Language Models (LLMs) has transformed the landscape of natural language processing, empowering applications ranging from chatbots and dialogue systems~\cite{zhang2019dialogpt} to content generation~\cite{radford2018improving, radford2019language}. 
However, as these models become an integral part of our communication, there are concerns about the potential biases (e.g., hallucinations~\footnote{The term ``hallucinations'' refers to factual errors~\cite{azaria2023internal}.}, toxicity, stereotypes) embedded within their outputs~\cite{deshpande2023toxicity, liu2023trustworthy, venkit2023unmasking, yang2023harnessing, zhuo2023red}. 
These subtle and implicit biases can reinforce stereotypes, marginalize certain groups, and perpetuate inequalities~\cite{birhane2023hate, henderson2018ethical}. 
Auditing LLMs for bias is thus essential for upholding ethical standards, reducing harm, and an inclusive deployment.

Despite advances in bias detection and mitigation strategies for LLMs in recent years, a large corpus of prior work has focused on word-level representations~\cite{basta2021extensive, bordia2019identifying, tan2019assessing}. 
However, in recent years, sentence-level representations from models such as GPT~\cite{radford2018improving, radford2019language} have become prevalent for text sequence encoding. 
Moreover, prior research has primarily operated under the assumption that the particular type of bias, such as hallucinations, is known \emph{a priori}, i.e., during the training phase~\cite{azaria2023internal, liang2020towards}. 
However, this assumption can potentially restrict these methods' practical applicability and generalizability since they rely on access to labeled data, which may necessitate a resource-intensive data collection process. 
Gaining access to labeled data can be particularly challenging when the occurrence of a bias is infrequent and given that language constantly evolves, which inherently leads to the continuous change of biases and their linguistic expressions.
Moreover, exposure to harmful content can lead to emotional and psychological stress for content moderators/labelers~\cite{miceli2022data}.

In this work, we propose an auditing approach to bias detection in LLMs sentence embeddings when the bias is \emph{not known a-priori}. 
Our goal is to determine if a pre-trained LLM has internalized harmful anomalous patterns (e.g., hallucinations) by examining its internal states (node activations). 
Following prior work~\cite{azaria2023internal, jamali2023unveiling}, our underlying assumption is that an LLM capable of predicting or generating anomalous content will exhibit detectable indicators of this tendency within its internal states.

Our method extends prior work on anomalous subset scanning for neural networks~\cite{cintas2021detecting,  kim2022out, ijcai2023p0107} by scanning pre-trained LLM activations.
Our method operates without the need for training data containing labeled anomalous content (e.g., false/hallucinated statements).
Unlike classifiers, we do not require a training phase for a particular bias type.
Instead, during testing, we rely on a reference dataset assumed to contain ``normal'' (or safe) content devoid of anomalies. 
When testing, we input a test dataset, possibly containing bias, to a pre-trained LLM model under audit. Our method examines the LLM's hidden states (activations) and identifies a subset of input data (e.g., sentences and nodes) as anomalous. 
We extend the prior work~\cite{cintas2021detecting,  kim2022out, ijcai2023p0107} to address that LLM embeddings for anomalous sentences may deviate in either direction from the expected distribution.
If anomalous sentences, e.g., those containing stereotypes, are detected from the activations, it suggests the model encodes those anomalous patterns. 
Conversely, the absence of anomalies detected in the activations of anomalous sentences may indicate the model's potential robustness to those patterns.\footnote{Note that we cannot confirm a null hypothesis (absence of anomalies).}

\subsection{Related Work}\label{sec:related-work}

\paragraph{Auditing LLM Outputs.}  
Prior work on detecting anomalies such as stereotypes, toxicity, or hallucinations in LLM models has concentrated on analyzing the model's generated content such as the percentage of anomalous options preferred or chosen~\cite{ kurita2019measuring, nangia2020crows}.
Other work has explored the propagation of bias to downstream tasks, including coreference resolution \cite{zhao2019gender}, sentiment analysis \cite{venkit2023unmasking}, topic modeling \cite{hassan2021unpacking}, and prediction models \cite{gehman2020realtoxicityprompts}. 
However, the effectiveness of these approaches is heavily reliant on the quality of pre-trained downstream models.
A different line of work has examined bias in the activations of LLMs, using principal component analysis \cite{basta2019evaluating, liang2020towards, zhao2019gender}, clustering \cite{basta2019evaluating}, or training detection classifiers on the latent space \cite{azaria2023internal, basta2019evaluating, bolukbasi2016man, henlein2022toothbrushes}. 
Other work has studied distance metrics between word pair representations \cite{ basta2019evaluating, bolukbasi2016man}. However, this approach has shown inconsistency detection results within contextual scenarios~\cite{henlein2022toothbrushes, kurita2019measuring, may2019measuring}. 
Furthermore, these approaches assume the availability of fully labeled training data and require predefined anomalous patterns. 
Few prior work has addressed the identification of unknown biases in LLMs, particularly in the context of unbiased sentence classification \cite{utama2020towards}.
In this work, our goal is to detect whether an LLM encodes anomalies (e.g., hallucinations) within its hidden states. We work under the assumption that only ``normal'' (e.g., true) data is available, while the presence of anomalous (e.g., false) data remains undisclosed.

\paragraph{DeepScan} In the context of analyzing data using a pre-trained network, deep subset scanning (DeepScan) ~\cite{cintas2021detecting} has been used to detect anomalous samples in various computer vision and audio tasks, including creativity sample characterization~\cite{cintas2022towards}, audio adversarial attacks in inner layers of autoencoders~\cite{akinwande2020identifying}, patch-based attacks in flow networks~\cite{ijcai2023p0107} and skin condition classification~\cite{kim2022out}. 
In this work, we extend prior work by scanning pre-trained LLM activations and introducing two novel methods to effectively identify anomalous sentences deviating from the expected activation distribution in either direction.

\section{DeepScan for LLMs}\label{sec:method}

This section introduces adaptations and extensions to deep subset scanning (DeepScan)~\cite{cintas2021detecting,ijcai2023p0107} for auditing LLM activations. 
Specifically, we detail the adaptation of previous deep scanning approaches to search for the ``most anomalous'' subset of node activations and input sentences within the inner layers of a pre-trained LLM network. 
We assume two datasets: a reference dataset $\Bcal$ containing $B$ ``normal'' (e.g., factually true) sentences, and an independent test dataset $\Tcal$ containing $M$ sentences, which may be either ``normal'' or ``anomalous'' (e.g., factually false). A sentence can represent any continuous text span and is not limited to a traditional linguistic sentence~\cite{devlin2018bert}.
\paragraph{Problem Formulation.} 
Consider an LLM, such as BERT \cite{devlin2018bert}, which can generate activations through its encoder\footnote{Or decoder for decoder-only models like GPT \cite{radford2018improving, radford2019language} or OPT~\cite{zhang2022opt}.}. 
Assume we have $M$ test sentences represented by a vector of activations $Z^l = [Z^l_1, \dots, Z^l_M]$ generated by the LLM at layer $l$, with each sentence activation having dimension~$J$ corresponding to the set of nodes $O^l = \{O^l_1, \dots, O^l_{J}\}$. 
Now, let $Z_S \subseteq Z$ and $O_S \subseteq O$, then we define a subset over sentences and nodes as $S = Z_S \times O_S$. 
Our goal is to identify the subset of activations containing the most anomalous (e.g., hallucinated) content based on a scoring function $F(S)$ that yields the anomaly score of a subset $S$: $S^\star=\arg \max _{S} F(S)$.

Detecting anomalies in activations typically requires parametric assumptions for the scoring function (e.g., Gaussian, Poisson). 
However, given that the distribution of activations in specific layers can be highly skewed, we adopt a non-parametric approach, following prior work \cite{chen2014non, cintas2021detecting, kim2022out,  mcfowland2013subsetscan, mcfowland2018efficient}. 
This approach, known as non-parametric scan statistics (NPSS), makes minimal assumptions about the underlying distribution of node activations. 
For this, we first derive $p$-values from the activations and then perform a scan over these $p$-values to quantify the difference or shift in the activation distribution for each dimension (node) compared to the reference distribution. We detail this now.

\paragraph{Empirical $p$-values.} 
In line with prior work~\cite{cintas2021detecting}, we utilize the activations from the (``normal'') reference data $\Bcal$ to compute empirical $p$-values for the activations from the (``normal'' or ``anomalous'') test data $\Tcal$.
For a given test activation $z^{\Tcal l}_{mj}$ (corresponding to sentence $m$, layer $l$ and node $j$), we calculate its empirical $p$-value by first sorting the set of activations from the reference data $\{z^{\Bcal l}_{bj}\}_{b=1}^B$
corresponding to layer $l$ and node $j$ across all reference sentences $b = 1 \dots B$ and then determining the rank of the test activation within that sorted list of reference activations. Subsequently, we normalize these positions to generate $p$-values within $[0, 1]$:
\begin{align}\label{eq:p-empirical-embeddings}
    p_{mj}^l = \frac{1 + \sum_{b = 1}^{B}\mathbf{1} (z^{\Bcal l}_{bj} \geq z^{\Tcal l}_{mj})}{1 + B}.
\end{align}
Here, $\mathbf{1}(\cdot)$ is the indicator function.
For ties, we consider both the rank on the left ($\texttt{pmin}$) and right sides ($\texttt{pmax}$). 
This gives us a range $[\texttt{pmin}, \texttt{pmax}]$. 
To obtain a single $p$-value within the range, we perform uniform sampling.
We compute $p$-values for a given layer $l$ for each node $j$ and sentence $m$ of the test activations. 

There are different methods for computing these empirical $p$-values. 
In left-tail (right-tail) $p$-values, the focus is on extreme values on the left (right) side of the reference activation distribution, indicating smaller (larger) values compared to the reference dataset.
For two-tailed $p$-values, the focus is on extreme values on both (left and right) sides.
Prior research concentrated on one-tailed $p$-values~\cite{cintas2021detecting, kim2022out} on the left side of the activation distribution. 
However, in our case, we observe deviations on both sides of the distribution and introduce two novel methods to incorporate the extreme values from both ends, as we detail below.

\paragraph{Uniform Distribution of $p$-values Under Null Hypothesis}
When the null hypothesis is true, it implies that any observed data point has an equal chance of falling anywhere within the distribution of possible values under the null hypothesis~\cite{murdoch2008p, rice2006mathematical}. 
Therefore, when we calculate empirical $p$-values by determining how extreme our observed data is relative to this null distribution, each potential outcome is equally likely. 
This uniformity in probabilities across the distribution ensures that $p$-values for samples confirming the null hypothesis follow a uniform distribution, as they are essentially measuring the randomness of the data in a manner consistent with the null hypothesis's assumptions.
Thus if the test dataset were to contain only ``normal'' sentences (null hypothesis), the $p$-values for test activations at layer $l$ would exhibit for each node $j$ a uniform distribution across sentences.
When anomalous sentences are introduced, and the LLM activations encode these anomalies, we hypothesize a departure from this uniform $p$-value distribution, particularly for certain nodes $j$.

\paragraph{Scoring Function.} 
To test whether the $p$-value distributions diverge from a uniform distribution, we employ a scoring function based on a test statistic, denoted as ${F(S) = \max_\alpha F_\alpha(S)}$, where $\alpha$ represents a significance level, and $F_\alpha(S)$ is defined by a suitable goodness-of-fit statistic. 

In the following explanation, we closely follow~\cite{cintas2021detecting}.
The general form of the scoring function is:
\begin{align}\label{eq:scoring-function}
\begin{split}
       F (S) & = \max_\alpha F_\alpha(S)  = \max_\alpha \phi (\alpha, N_\alpha(S), N(S))
\end{split}
\end{align}
where $N(S)$ represents the number of empirical p-values contained in subset $S$, $N_{\alpha}(S)$ is the number of $p$-values less than (significance level) $\alpha$ contained in subset $S$,  $\alpha \in (0,1)$ is a significance level and $\phi$ is a goodness-of-fit statistics. 
To identify a subset $S$ that presents the strongest indication of significantly exceeding the expected activation distribution under the null hypothesis (``normal'' or clean data). 
This is expressed by the condition $N_{\alpha}(S) > \alpha N(S)$, where $\alpha$ denotes the chosen significance level. 
In our experiments, we run a grid search over $\alpha \in[0.05, 0.5]$ in steps of $0.05$.

\paragraph{Higher Criticism Test Statistic}
While there are several established goodness-of-fit statistics available for use in NPSS \cite{mcfowland2018efficient}, in this work, we utilize the Higher Criticism (HC) test statistic \cite{donoho2004higher}:
\begin{equation}
    \phi (\alpha, N_\alpha(S), N(S)) = \frac{|N_\alpha(S) - N(S)\alpha|} {\sqrt{N(S)\alpha(1-\alpha)}}
\end{equation}
This could be understood as the test statistic for a Wald test assessing the number of significant $p$-values, where $N_{\alpha}$ follows a binomial distribution with parameters $N_{\alpha}$ and $\alpha$.
Due to its normalization by the standard deviation of $N_{\alpha}$, HC tends to yield smaller subsets characterized by wider ranges of $p$-values. 
This occurs because such subsets yield larger values in the numerator while generating smaller values in the denominator.
In our case, small subsets are particularly preferable in scenarios where the quantity of anomalous data within the test dataset is small, as shown in our experiments (Section~\ref{sec:experiments-audits}), where the test dataset comprises only 10-20\% anomalous data.

\paragraph{Efficient Search Algorithm.}
To overcome the computational challenge posed by maximizing a scoring function across all possible data sample and node subsets, we employ Fast Generalized Subset Scanning (FGSS) as previously utilized in similar research \cite{cintas2021detecting, kim2022out}. 
This approach significantly reduces the number of subsets under consideration from $O(2^E)$ to $O(E)$ within each optimization step, where $E$ represents the number of elements currently being optimized, such as images or nodes (see Appendix~\ref{apx:setup},  Algorithm~\ref{alg:optimize_rows}). 
This efficiency is based on the application of the LTSS property \cite{neill-ltss-2012}, which involves sorting each element based on its priority, defined as the proportion of $p$-values below a threshold $\alpha$. 
FGSS assures convergence to a local optimum.
The algorithm returns anomalous subset $S^{\star}$ defined by a set of nodes $O_{S^{\star}}$ and a subset of sentences $Z_{S^{\star}}$ from the test dataset, collectively defining the most anomalous pattern as a group.

\paragraph{Using results from both tails.}
We observe that LLM embeddings for anomalous data may shift from the expected reference distribution in both directions. 
To identify subsets marked as anomalous due to shifts to the left or right for different nodes, we introduce two novel methods to aggregate scanning results.
The first approach involves aggregating results obtained from scanning left-tail and right-tail $p$-values. We identify subset of sentences $Z_{S^{\star}_R}$ by scanning over right-tailed $p$-values and $Z_{S^{\star}_L}$ by scanning over left-tailed $p$-values.
We then combine these results through union: $Z^{\mathrm{Union}}_{S^{\star}}: = Z_{S^{\star}_R} \cup Z_{S^{\star}_L}$.
The second approach is an iterative method that combines results from scanning two-sided $p$-values. It aggregates the top-$k$ subsets returned by the scanning, where after each iteration $i = 1 \dots k$, the found subset of sentences $Z_{S^{\star}_i}$ is removed from the test dataset, and the subsequent scanning is performed on the reduced test set. 
The final subset is the union over all identified subsets: $Z_{S^{\star}}^{\mathrm{top-}k}: = Z_{S^{\star}_1} \cup Z_{S^{\star}_2} \cup \dots Z_{S^{\star}_k}$.

\section{Experimental Setup and Results}\label{sec:experiments-audits}
This section presents experimental results for bias detection in LLMs using our two proposed scanning methods. 
We focus on hallucination detection and analyze the subset of input sentences returned. For additional results on toxicity and stereotype detection, refer to Appendix~\ref{apx:addtional-results}.

\paragraph{Data and LLM Models.}
We use an English-language dataset \hallucinations~\cite{azaria2023internal} ('Cities' topic) containing factually true (e.g., ``Nakuru is a city in Kenya.'') and false statements (e.g., ``Surrey is a city in Kenya.'').  
We use a test dataset of $800$ samples comprising $10$\% anomalous data, reflecting the real-world scarcity of such data. 
We sample test data $10$ times with replacement from a larger pool of data and report mean and standard deviation.
For further details regarding the dataset selection and preprocessing, see Appendix~\ref{apx:data}. 
We audit two pre-trained LLMs: the BERT base (uncased) model~\cite{bertbaseuncased} with a $12$-layer encoder with $768$ nodes per layer, where focus on activations for the \cls token, and the Facebook OPT 6.7 model~\cite{zhang2022opt} with a $32$-layer decoder with $4096$ nodes per layer.
Note that our method can audit any LLM that provides activations.
For details, see Appendix~\ref{apx:models}.

\paragraph{DeepScan Extensions and Baseline.} 
We scan over left-, right-, and two-tailed $p$-values. Building upon prior work~\cite{azaria2023internal}, we analyze the activations from layers closer to the output as they are suspected of encoding higher-level information. 
For further details, see Appendix~\ref{apx:setup}.
We report results for the introduced scanning extensions: the outcomes derived from the combined subset of left and right $p$-value scans (\scanlr), along with the results from the top-$3$ scan over the two-sided $p$-values (\scantwo).
To evaluate our detection power, we compare to a supervised classifier (\clfshift) that aims to predict whether a sentence is true or false based on the LLM activation~\cite{azaria2023internal}. 
It is trained for each LLM layer on an approximately balanced dataset containing $1739$ false sentences and is tested on an out-of-distribution held-out dataset (e.g., other topics) of the same task.
For details, see Appendix~\ref{apx:experiments}.

\paragraph{Results.}
 \begin{table}
     \centering
     \begin{tabular}{lrrllll}
\toprule
 LLM &  Layer &    Clf &     Precision &        Recall &          Size \\
\midrule
\multirow{6}{*}{BERT} &    \multirow{3}{*}{10}  &   clf+ & \B 0.099 (0.006) & \B  0.73 (0.046) &  0.737 (0.01) \\
 &      &  \scantwo & \B 0.114 (0.022) & 0.631 (0.105) & \B 0.563 (0.075) \\
 &      & \scanlr & 0.091 (0.017) &  0.61 (0.062) & 0.685 (0.075) \\
 \cline{2-6} 
 &     \multirow{3}{*}{12}  &   clf+ & \B 0.128 (0.017) & \B 0.428 (0.059) & \B 0.335 (0.012) \\
 &       &  \scantwo & 0.085 (0.011) & \B 0.392 (0.062) &  0.459 (0.018) \\
 &       & \scanlr & 0.065 (0.029) & \B 0.465 (0.246) & 0.695 (0.074) \\
 \hline
\multirow{6}{*}{OPT}  &       \multirow{3}{*}{20}   &   clf+ &  0.33 (0.018) & \B 0.742 (0.041) &  0.225 (0.01) \\
  &      &  \scantwo &  0.16 (0.057) &   \B 0.734 (0.2) & 0.479 (0.065) \\
  &      & \scanlr &  \B 0.45 (0.034) & 0.605 (0.055) &  \B 0.134 (0.01) \\
   \cline{2-6} 
  &       \multirow{3}{*}{24}   &   clf+ & 0.274 (0.015) & \B 0.752 (0.034) & 0.275 (0.013) \\
  &     &  \scantwo & 0.149 (0.031) & 0.685 (0.096) & 0.471 (0.094) \\
  &     & \scanlr & \B 0.693 (0.272) & 0.418 (0.146) & \B 0.076 (0.042) \\
\bottomrule
\end{tabular}

     \vspace{2pt}
     \caption{Comparison of our weakly supervised scan methods unions: left- and right-tailed  $p$-value scans (\scanlr), and top-$3$ two-tailed $p$-value scans (\scantwo), and the supervised out-of-distribution classifier baseline  (\clfshift) on auditing BERT and OPT.
     Performance and relative subset (Size) reported as mean~(std) across $10$ random test datasets with $10$\% anomalous data. Best (significant) bold. }
     \label{tab:comparison-scan-clf}
 \end{table}
We present results in Table~\ref{tab:comparison-scan-clf}.
After identifying the subset, we assume access to test labels and report precision, and recall.
Precision measures the ability to avoid false positives, calculated as the ratio of correctly identified anomalous samples to the total samples flagged as anomalous. 
Recall quantifies the ability to find all anomalous sentences, measured as the ratio of correctly identified anomalous samples to the total actual anomalous instances.
We also report the size of the subgroup of sentences returned by the scanner or the number of sentences predicted as false by the classifier.

We first audit BERT. Across activations from both layer $10$ and (last) layer $12$, we observe low precision across all methods.
These results indicate that BERT has a limited capacity to represent hallucinations within its internal state effectively, confirming prior work~\cite{azaria2023internal}.
Subsequently, we audit OPT, a more potent model designed to match GPT's capabilities~\cite{zhang2022opt}. 
We consistently observe higher precision in detecting false statements across methods and test sets. 
From an auditing perspective, these findings indicate that OPT does indeed encode hallucination information within its internal state.
This is consistent with prior research~\cite{azaria2023internal}, which found layer 20 to be the most predictive for their classifier.
Yet, our scanning approach (\scanlr) excels at layer 24.

Comparing methods, \scantwo\ achieves similar precision levels to the classifier for BERT. At layer $10$, the classifier shows highest recall while flagging $\sim 74$\% of sentences as false, despite an expected rate of $10$\%, indicating a high False Positive Rate.
In the case of OPT, \scanlr\ exhibits higher precision than the supervised baseline for layer $20$, with a subset size ($\sim$ $13$\%) closely matching the expected $10$\%. 
Nevertheless, \scantwo\ and \clfshift\ achieve the highest recall rates and return larger subsets for both layers.
In summary, method effectiveness varies with test dataset and layers, without a clear dominant method. 
Rather, we observe a trade-off between precision and recall, with one method excelling in precision at the cost of recall, and vice versa.
We also observe a strong connection between subset size and recall, where larger subsets tend to yield higher recall but often at the cost of decreased precision.
In conclusion, our method---with no prior exposure to false statements---exhibits similar performance to an out-of-distribution classifier trained on larger amounts of anomalous samples when both are assessed using a dataset comprising just $10$\% anomalous data ($80$ samples).
\section{Summary and Discussion}\label{sec:summary}
We have introduced a weakly supervised auditing technique to identify, whether a pre-trained LLM is encoding patterns such as hallucinations within its internal states. 
We are interested in this problem because if an LLM is encoding these patterns internally, it can potentially impact downstream tasks and one may be able to deploy bias mitigation strategies. 
Our method employs subset scanning across various neural network layers in pre-trained LLMs, without the need for for prior knowledge of the specific patterns or access to labeled false statements.

During validation on a hallucination dataset, our approach achieved performance similar to, and sometimes surpassing a baseline fully supervised out-of-distribution classifier. 
Importantly, our approach only requires access to samples labeled as ``normal'' (true) eliminating the need for anomalous pattern data, which can be costly and ethically challenging to obtain (especially for other types of anomalies, such as toxicity and stereotypes). 
This makes our method highly suitable for real-world applications.
Nonetheless, recent research~\cite{levinstein2023still} has raised doubts about the generalizability of prior methods~\cite{azaria2023internal, burns2022discovering} in detecting hallucinations, which we intend to explore further.
Our work makes assumptions about the background dataset, assuming it \emph{only} contains ``normal'' statements tailored to the problem at hand. We plan to extend our work to more realistic assumptions by including small amounts of anomalous data in the reference dataset and composing it of various data sources.

Finally, similar to work using metrics such as cosine-similarity~\cite{liang2020towards}, our method only has positive predictive ability: it can be used to detect the presence of anomalies but not their absence. 
To understand how much of our experimental results can be attributed to our method's detection power and how much is due to the LLM not encoding the anomalous pattern, we have compared it to a supervised baseline. 
Our results indeed show that our method, which requires no training and no prior exposure to false statements, performs comparably to the supervised baseline.
\section{Outlook: Informing Fine-tuning}\label{sec:outlook-tuning}
We briefly discuss the potential expansion of our work. 
Unsupervised pre-training and task-specific fine-tuning have become the standard approach for various LLM tasks~\cite{hadi2023survey, huang2023chatgpt, thirunavukarasu2023large}.
Despite their impressive achievements, these methods face challenges in generalization performance on downstream tasks~\cite{devlin2018bert, lee2020mixout, phang2018sentence} and suffer from catastrophic forgetting~\cite{aghajanyan2020better, mahabadi2021variational, xu2021raise}.
To address these issues, sub-network optimization approaches have emerged as a promising method to enhance stability and reduce overfitting without requiring full network retraining~\cite{tong2023bi,xu2021raise}.

We believe that our method can build on the advancement of this line of research for bias mitigation strategies. 
As detailed in \S~\ref{sec:method}, our method allows identifying the subset of nodes $O_{S^{\star}}$ that are most responsible for the anomalous patterns found.
This means, for each layer, we are able to identify the nodes that align with the most anomalous subset of sentences, that is those nodes for which the empirical $p$-values of their activations deviate from the uniform distribution such that they are flagged anomalous. 
This suggests that these nodes are pivotal in identifying anomalous patterns within the data, which could guide the efficient fine-tuning of sub-networks for bias mitigation strategies.
We show initial findings in Appendix~\ref{apx:addtional-results}.

\section*{Acknowledgements}
We appreciate Amos Azaria and Tom Mitchell for sharing their data and code. Special thanks to Edward McFowland III for valuable feedback and discussions.

\newpage
\bibliography{refs}

\newpage
\appendix

\section{Datasets}\label{apx:data}
This section provides additional information on the datasets used and their curation for our use.

\subsection{\hallucinations\ Dataset} The \hallucinations\ dataset~\cite{azaria2023internal}\footnote{The authors generously provided us with access to the dataset upon our request.} provides a collection of true and false statements across six domains: Cities, Inventions, Chemical Elements, Animals, Companies, and Scientific Facts. 
The dataset has been curated by the authors using reliable sources utilized to craft true statements while corresponding false statements are generated using distinct values from the same source. 
The dataset maintains the same amount of true and false statements for each topic. 
The authors express their intention to release the dataset publicly and have kindly granted us access in advance. 
The dataset consists of a total of six topics: Cities ($7573$ samples after deleting duplicates), Inventions ($876$), (Chemical) Elements ($930$), Animals ($1008$), Companies ($1200$), and (Scientific) Facts ($612$). 
We use the Cities dataset as test data. Importantly, when we mention ``hallucinations'' data hereinafter, we assume it pertains to the Cities topic unless specified otherwise.
We evenly divide the true statements into the reference and clean batches while utilizing the false statements as the anomalous batch.

\subsection{\realtoxicity\ Dataset}
The \realtoxicity\ dataset~\cite{gehman2020realtoxicityprompts} comprises 100,000 sentence snippets sourced from the web. 
Each instance includes a prompt and a continuation. 
The prompt includes attributes such as text, profanity, sexually explicit content, identity attack, flirtation, threat, insult, severe toxicity, and toxicity scores. The dataset is structured to offer insights into various aspects of toxicity. 
The scores accompanying the prompt and continuation are calculated using the Perspective API. 
The prompts were selected from sentences in the OPEN-WEBTEXT CORPUS, with toxicity scores ($[0, 1]$) extracted using the PERSPECTIVE API\footnote{\url{https://github.com/conversationai/perspectiveapi}}. 
The dataset was curated to ensure a stratified representation of prompt toxicity across different ranges. 

We rely on a test dataset provided by HuggingFace\footnote{\url{https://huggingface.co/datasets/allenai/real-toxicity-prompts}}, comprising $99442$ samples. 
We concatenate the prompt and the continuation, subsequently deriving the toxicity score by calculating the average of the scores from both sentences. 
If a score is not available for a continuation, we rely on the toxicity score from the prompt.
We first split the dataset into harmless and toxic sentences. As prior work~\cite{wang2021simple}, we assume sentences with a toxicity score below $0.3$ as harmless scores and sentences with a toxicity score above $0.7$. 
We split the dataset into three non-overlapping batches: reference ($22479$ samples), clean ($22479$ samples), and anomalous ($3642$ samples). 
For the experiments, we reduced the reference and clean sample size to $2000$ sentences.

\subsection{\stereoset\ Dataset}
The \stereoset~\cite{nadeem2020stereoset} dataset is designed to assess stereotype bias in language models and comprises $17000$ sentences that evaluate model preferences across gender, race, religion, and profession.
It supports tasks like multiple-choice question answering.
The dataset is available in English and is structured into \texttt{intersentence} and \texttt{intrasentence} categories. 
Each data instance contains information such as the bias type (e.g., gender, race), context sentence, and target sentence. 
The dataset also includes gold labels indicating whether sentences are \texttt{stereotypical}, \texttt{anti-stereotypical}, or \texttt{unrelated}. 
The data was collected through crowdworkers, with annotations provided by $475$ and $803$ annotators for {intrasentence} and {intersentence} tasks, respectively.
The dataset's focus on measuring bias within various domains underscores its relevance for evaluating language model fairness and comprehension. 
Note that the stereotypes are grounded in US contexts, and some sentences generated by annotators may be objectively false or include favorable stereotypes. 
Recent work emphasizes the importance of community engagement to enhance dataset curation strategies, addressing challenges in achieving comprehensive representation across diverse global cultures and perspectives when building stereotype repositories~\cite{dev2023building}.

We rely on a validation dataset provided by HuggingFace\footnote{\url{https://huggingface.co/datasets/stereoset}}, comprising $2123$ samples for intersentence, and $2106$ samples for intrasentence. 
Our approach involves random sampling from our dataset to form reference, clean, and anomalous batches. 
Each batch comprises $702$ samples, while we ensure that sentences do not overlap within each batch. 
In this context, we define \texttt{unrelated} as representing the normal data, while \texttt{stereotypical} signifies the anomalous pattern. 
For the \texttt{intrasentence} scenario, we select the appropriate sentence from the dataset (\texttt{unrelated} or \texttt{stereotypical}). 
For the \texttt{intersentence} scenario, we generate sentence samples by combining the context sentence with the corresponding sentence continuation (\texttt{unrelated} or \texttt{stereotypical}).

\section{LLM Models}\label{apx:models}
%
This section provides additional details on the large language models (LLMs) audited in our experiments.

\paragraph{BERT base}
We use the BERT base model (uncased)~\cite{bertbaseuncased} provided by HuggingFace\footnote{\url{https://huggingface.co/bert-base-uncased}}, which is a pre-trained transformer model designed for the English language, and the corresponding Tokenizer. 
BERT is trained via self-supervised training on unlabeled raw texts, utilizing two objectives: masked language modeling and next-sentence prediction. 
Its training data includes a large corpus of English texts, and it is important to note that the model can have biased predictions. 
The model has been primarily designed for fine-tuning downstream tasks such as sequence classification and token classification.

BERT is constructed as a transformer-based model, which employs an encoder architecture with a self-attention mechanism that allows it to capture contextual information from both preceding and following words in a sentence, creating a bidirectional understanding. 
The BERT base model encoder has $12$ layers, and $12$ attention heads per layer. The embedded space of the model has a dimension of $768$.
In BERT, the \cls token is a special token added to the beginning of each input sequence. 
The final hidden state of the \cls token is often used as a summary representation of the entire input sequence for downstream classification tasks. 
Our analysis focuses on the hidden representations of the \cls token, as it is commonly used for downstream classification tasks~\cite{kim2021self, reimers2019SBERT}.

\paragraph{OPT}
We use the Open Pre-trained Transformer (OPT) language model~\cite{zhang2022opt} ($6.7$B) provided by HuggingFace\footnote{\url{https://huggingface.co/facebook/opt-6.7b}}. 
It has been specifically trained to align with the performance and sizes of the GPT-$3$ models employing causal language modeling (CLM) objectives. 
The training data for OPT comprises a fusion of filtered datasets sourced from various origins, such as BookCorpus, CC-Stories, The Pile, Reddit, and CCNewsV2, which is comprised mainly of English text, with a small quantity of non-English data. 
OPT serves multiple purposes, including text generation and evaluation via prompts, and can be fine-tuned for specific tasks. 
The training of the pre-trained model spanned approximately $33$ days and harnessed the power of multiple GPUs. 
Note that the model is susceptible to biases and safety concerns due to its training on a diverse and unfiltered array of internet data. 

The model utilizes a decoder-only pre-trained transformer architecture. 
It consists of $32$ layers, where each layer is composed of $4096$ hidden units. 
For a data and model card, see~\cite{zhang2022opt}.

\section{Scanner Details and Setup}\label{apx:setup}
%
This section provides additional details on the setup of our scanning approach. 
For more details see prior work~\cite{cintas2021detecting, kim2022out}.

\paragraph{Visual Exploration of Activations and Empirical $p$-values}
In \S~\ref{sec:method}, we stated that to detect anomalies in activations we would typically need to rely on parametric assumptions, but due to the potential skewness in specific layer activations, we prefer to rely on a non-parametric approach with minimal assumptions about the underlying distribution of activations. 
Instead, we detect anomalies in $p$-values derived from activations. 
We expect the $p$-values from clean samples to be uniformly distributed, while we expect the $p$-values from anomalous samples not to be distributed uniformly.

\begin{figure}
     \centering
     \begin{subfigure}[b]{0.24\textwidth}
         \centering
         \includegraphics[width=\textwidth]{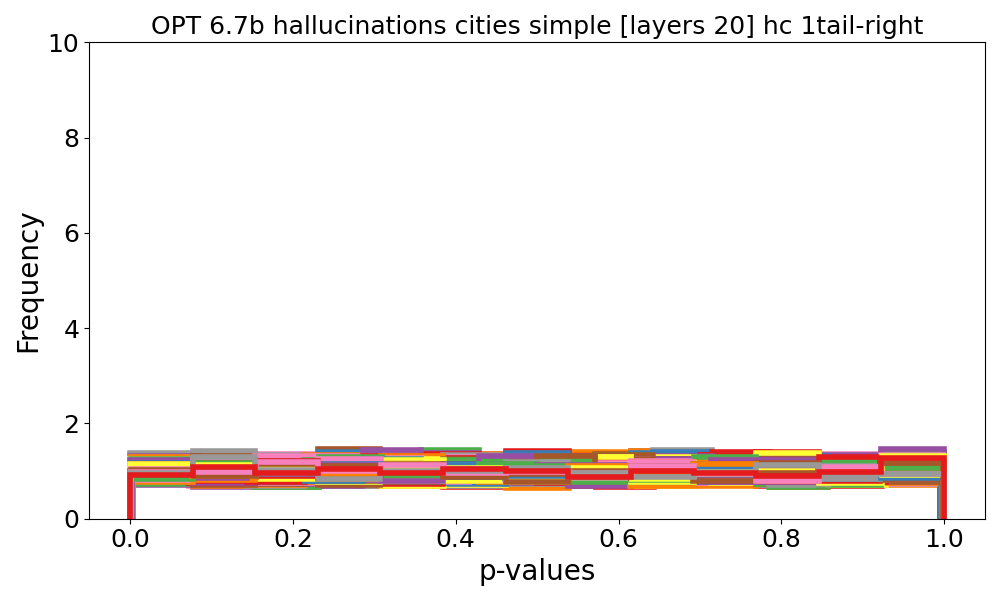}
         \caption{$p$-values (clean).}
         \label{fig:pvalues-clean}
     \end{subfigure}
     \hspace{2cm}
     \begin{subfigure}[b]{0.24\textwidth}
         \centering
         \includegraphics[width=\textwidth]{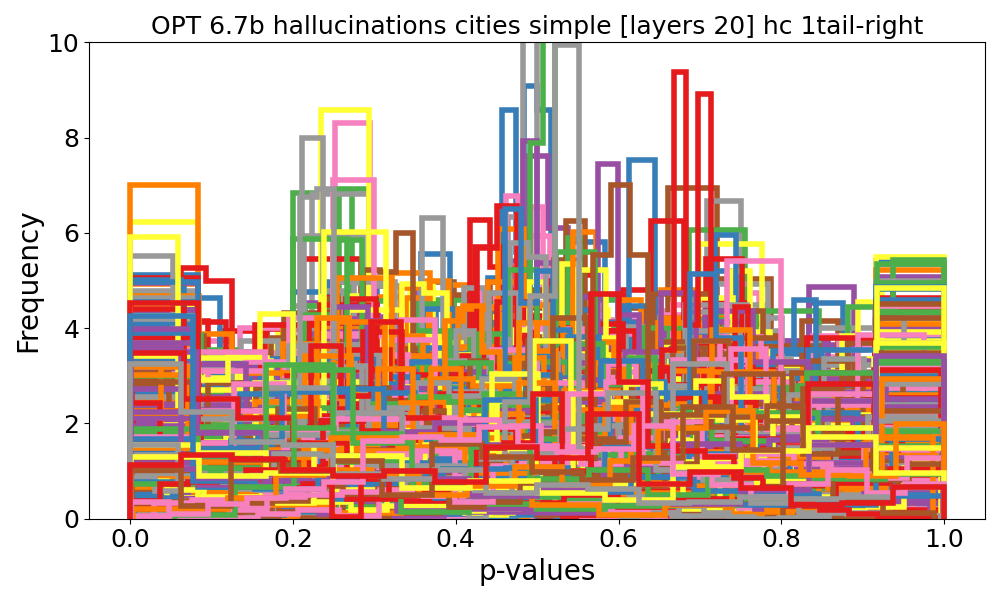}
         \caption{$p$-values (anom).}
         \label{fig:pvalues-anom}
     \end{subfigure}
     \caption{Distributions right-tailed empirical $p$-values for activations from all $4096$ nodes (colors) of OPT 6.7b layer 20 over test samples from \hallucinations\ data. Clean samples refer to true statements, and anomalous (anom) samples refer to false statements.}
     \label{fig:activations-pvalues}
\end{figure}

We provide a visual exploration of this concept in Figure~\ref{fig:activations-pvalues}. 
Specifically, we study the $p$-values derived from activations from OPT 6.7b activations in layer $20$, using the hallucination dataset. Each node's distribution is depicted in a distinct color.
We examine the distribution of empirical $p$-values computed under the reference dataset (containing ``clean'') statements for clean (Fig.~\ref{fig:pvalues-clean}) and anomalous (Fig.~\ref{fig:pvalues-anom}) data. We observe that the $p$-values under the null hypothesis (``clean'' data) exhibit a uniform distribution, whereas the $p$-values for the anomalous data significantly deviate from uniformity. 
Our method leverages this difference in $p$-value distributions to identify anomalous patterns.

\paragraph{Berk Jones Test Statistic}
In Appendix~\ref{apx:addtional-results}, we also report results that use the Berk-Jones (BJ) test statistic ~\cite{berk1979goodness}:
\begin{align}\label{eq:bj-test}
\begin{split}
        \phi (\alpha, N_\alpha(S), N(S)) = N(S) KL\left( \frac{N_{\alpha}(S)}{N(S)}, \alpha \right),
\end{split}
\end{align}
where $KL$ is the Kullback-Leibler divergence $\operatorname{KL}(x, y) = x \log \frac{x}{y} + (1-x) \log \frac{1-x}{1-y}$ between the observed and expected proportions of significant $p$-values.

\paragraph{Fast Generalized Subset Scanning Algorithm}
We follow prior work~\cite{cintas2021detecting, kim2022out} in employing an iterative optimization process to identify anomalous subsets of $p$-values (see Algorithm~\ref{alg:singlerestart} and Algorithm~\ref{alg:optimize_rows}). 

\begin{algorithm}
\caption{Single Restart over $M$ test sentences and $J$ nodes}
\label{alg:singlerestart}
\begin{algorithmic}[1]
\Procedure{SingleRestart}{$(M \times J)$ $p$-values}
\State score $\leftarrow -1$
\State $Z_s \leftarrow$ \Call{Random}{$M$}
\State $O_s \leftarrow$ \Call{Random}{$J$}
\While{score is increasing}
    \State $(M \times J') = (M \times J) | O_s$
    \State score, $Z_s \leftarrow$ \Call{OptimizeRows}{($M \times J'$)}
    \State $(M' \times J) = (M \times J) | Z_s$
    \State score, $O_s \leftarrow$ \Call{OptimizeRows}{$(J \times M')$}
\EndWhile
\State \Return score, $Z_s$, $O_s$
\EndProcedure
\end{algorithmic}
\end{algorithm}

\begin{algorithm}
\caption{Optimize over rows using LTSS ({\em OptimizeRows}). It maintains maxscore and maxsubset over $\|E\|*\|T\|$ subsets.}
\label{alg:optimize_rows}
\begin{algorithmic}[1]
\Procedure{OptimizeRows}{$p$-values from all rows E and relevant cols C}
\State maxscore $\leftarrow -1$
\State maxsubset $\leftarrow \emptyset$
\For{$t$ in T = \Call{LinearSpace}{0,1}}
\State sortedpriority $\gets$ \Call{SortByPropCT}{E, t}
\State Score(sortedpriority, $t$)
\EndFor
\State \Return maxscore, maxsubset
\EndProcedure
\end{algorithmic}
\end{algorithm}

\section{Additional Information on Experiments}\label{apx:experiments}
%
In this section, we provide additional information on the experiments, especially reporting test, reference, and training dataset sizes. 
Finally, we also report details on the code and computational resources used to run experiments.
\subsection{Test Data for Scanner and Classifier}
For testing purposes, we select for \hallucinations\ and \realtoxicity\ datasets a total of $800$ samples from both clean and anomalous data at varying proportions as specified in our experiments. 
To illustrate, in the case of a $90:10$ split, we draw $90$\% of the $800$ samples from the clean dataset, and $10$\% of the $800$ samples from the anomalous dataset, also containing $2000$ samples.
For the \stereoset\ dataset, we chose a total of $400$ samples from clean and anomalous data (each containing a total of $500$ samples) across various mixing variations as described above.

\subsection{Reference Dataset for Scanner}
The reference, clean, and anomalous datasets contain $2000$ samples for \realtoxicity, $1500$ samples for \hallucinations, and $500$ for \stereoset. 

\subsection{Training Data for Classifier}
Following~\cite{azaria2023internal}, the baseline classifier is trained to predict whether a sentence is anomalous or not based on the LLM hidden layer's values. 
The classifier is composed of four fully connected layers featuring ReLU activation functions and a progressively decreasing number of neurons ($256$, $128$, $64$). 
The final layer employs a sigmoid activation function for binary classification. 
Training is conducted over $5$ epochs with a batch size of $32$, using the Adam optimizer and a $80:20$ training-validation split.

We train two types of classifiers classifier, one trained on the datasets from the same distribution as the test dataset (\clfstar) and one trained on a different distribution than the test dataset (\clfshift). 
We train \clfstar\ on \hallucinations\ (topic: Cities) using $989$ anomalous, and $989$ clean samples. 
For training \clfshift\ on \hallucinations\ (topics: Inventions, Elements, Animals, Facts) we use $1739$ anomalous, and $1688$ clean samples.

In Appendix~\ref{apx:addtional-results}, we provide additional results for \realtoxicity\ and \stereoset\ data. 
Due to the lack of a suitable out-of-distribution dataset, we report the results of in-distribution classifiers as a fully supervised best-method comparison. 
Specifically, we train \clfstar\ on \stereoset\ (inter-sentence, normal: unrelated, anomalous: stereotypes) using $208$ anomalous, and $415$ clean samples. 
We train \clfstar\ on \realtoxicity\ (normal: toxicity score in $[0, 0.3]$, anomalous: toxicity score  $[0.7, 1]$) using $1642$ anomalous, and $2000$ clean samples.

\subsection{Computational Resources and Code}
All experiments were conducted on a MacBook Pro (Apple M1 Max chip).
For our code, we rely on the publicly available GitHub repository: \url{https://github.com/Trusted-AI/adversarial-robustness-toolbox/tree/main/art/defences/detector/evasion/subsetscanning}.

\section{Additional Results}\label{apx:addtional-results}
In this section, we present the following additional results:
\begin{itemize}
    \item In \ref{apx:additional-toxicity}, we provide additional results for the \realtoxicity\ dataset.
    \item In \ref{apx:additional-stereoset}, we provide additional results for the \stereoset\ dataset.
    \item In \ref{apx:additional-testset}, we provide additional results for test datasets with larger amounts of anom. data.
    \item In \ref{apx:additional-scanlr}, we offer additional insights into our proposed \scanlr\ method.
    \item In \ref{apx:additional-scantwo}, we offer additional insights into our proposed \scantwo\ method.
    \item In \ref{apx:additional-outlook}, we explore the returned subset of nodes, as introduced in \S~\ref{sec:outlook-tuning}.
\end{itemize}

\paragraph{Metrics.} 
After identifying the subset, we assume access to test labels and calculate the following performance metrics

\begin{itemize}
    \item Precision measures the ability to avoid false positives, calculated as the ratio of correctly identified anomalous samples to the total samples flagged as anomalous.
    \item Recall quantifies the ability to find all anomalous sentences, measured as the ratio of correctly identified anomalous samples to the total actual anomalous instances.
    \item For classifiers, we also sometimes report accuracy reported by prior work~\cite{azaria2023internal}. Accuracy measures the overall correctness of the classification, calculated as the ratio of correctly classified samples---both anomalous and ``normal''---to the total number of test samples. 
\end{itemize}

Note that accuracy takes both true positives and true negatives into account, but it may not be as informative in highly imbalanced datasets where the number of ``normal'' instances significantly outweighs the anomalous ones (as in the case reported in the main paper).
In such cases, precision and recall are often more informative metrics.

Additionally, we include the measurement of the proportion of samples identified as anomalous or biased in relation to the size of the test set. 
As the subset size increases (i.e., an increase in both True Positives and False Positives), recall tends to increase as it becomes more likely that the model will correctly identify more of the positive samples.

\paragraph{Methods}

We compare the following methods:

\begin{itemize}
\item \textbf{\scanl}: Conducts a group scan over one-tailed left-sided empirical $p$-values obtained from activations.
\item \textbf{\scanr}: Conducts a group scan over one-tailed right-sided empirical $p$-values obtained from activations as prior work~\cite{cintas2021detecting, kim2022out}.
\item \textbf{\scanlr}: Our first proposed method performs a group scan that combines \scanl\ and \scanr\ and returns the union of the anomalous subset of sentences detected by each individual method (see \S~\ref{sec:method}).
\item \textbf{\scantwo}: Our second proposed novel method conducts a top-$k$ group scan over two-tailed empirical $p$-values obtained from activations and unions the found subsets (see \S~\ref{sec:method}). 
\item \textbf{\clfshift} is the out-of-distribution classifier proposed by prior work~\cite{azaria2023internal} 
\end{itemize}

\subsection{Additional Results on \realtoxicity}\label{apx:additional-toxicity}
In Table~\ref{tab:additional-toxicity} we report additional information on the performance of our two proposed scanning methods (\scanlr, \scantwo) on the \realtoxicity\ dataset described in Section~\ref{apx:data}. Note that due to the lack of out-of-distribution data, we do not report a baseline classifier (\clfshift).

\begin{table}
    \centering
    \begin{tabular}{llrllll}
\toprule
 LLM & Anom &  Layer &   Method &     Precision &        Recall &          Size \\
\midrule
\multirow{24}{*}{BERT}  &  \multirow{8}{*}{10\%} &     \multirow{4}{*}{10} & scan2 k1 & \B 0.121 (0.028) & 0.205 (0.058) & \B 0.169 (0.018) \\
 &  &     & scan2 k2 & \B  0.129 (0.02) & 0.442 (0.071) & 0.343 (0.014) \\
 &  &     & scan2 k3 & \B 0.127 (0.012) & \B  0.581 (0.046) & 0.458 (0.028) \\
 &  &     &   scanLR &  \B 0.12 (0.026) & 0.228 (0.067) & 0.188 (0.025) \\
           \cline{3-7} 
 &  &     \multirow{4}{*}{12} & scan2 k1 &  \B 0.12 (0.104) & 0.094 (0.059) & \B 0.106 (0.066) \\
 &  &     & scan2 k2 & \B 0.084 (0.021) &  0.244 (0.08) & 0.292 (0.078) \\
 &  &     & scan2 k3 & \B 0.091 (0.016) & \B  0.37 (0.061) & 0.411 (0.044) \\
 &  &     &   scanLR & \B 0.145 (0.128) & 0.095 (0.049) & \B 0.112 (0.067) \\
           \cline{2-7} 
 &   \multirow{8}{*}{20\%}  &     \multirow{4}{*}{10} & scan2 k1 & 0.238 (0.022) & 0.206 (0.028) & \B  0.173 (0.019) \\
 &  &     & scan2 k2 & \B 0.256 (0.023) & 0.451 (0.056) & 0.352 (0.017) \\
 &  &     & scan2 k3 & \B 0.259 (0.021) & \B 0.612 (0.062) & 0.473 (0.029) \\
 &  &     &   scanLR &  \B  0.27 (0.03) & 0.446 (0.199) & 0.322 (0.128) \\
           \cline{3-7} 
 &  &     \multirow{4}{*}{12} & scan2 k1 & 0.199 (0.117) & 0.084 (0.057) & \B 0.092 (0.065) \\
 &  &     & scan2 k2 & \B 0.168 (0.042) & 0.234 (0.082) &  \B 0.28 (0.079) \\
 &  &     & scan2 k3 & \B 0.192 (0.031) &\B  0.388 (0.088) & 0.402 (0.045) \\
 &  &     &   scanLR & \B 0.244 (0.144) & 0.199 (0.286) & \B 0.156 (0.185) \\
           \cline{2-7} 
 &   \multirow{8}{*}{50\%}  &     \multirow{4}{*}{10} & scan2 k1 &  0.57 (0.027) & 0.204 (0.019) & 0.179 (0.016) \\
 &  &     & scan2 k2 &  \B 0.61 (0.028) &  0.452 (0.03) &  \B 0.372 (0.03) \\
 &  &     & scan2 k3 & \B 0.611 (0.013) &  \B 0.599 (0.04) &  0.49 (0.034) \\
 &  &     &   scanLR & \B 0.641 (0.033) & 0.418 (0.171) & 0.324 (0.126) \\
           \cline{3-7} 
 &  &     \multirow{4}{*}{12} & scan2 k1 & 0.491 (0.046) & 0.089 (0.049) & 0.092 (0.051) \\
 &  &     & scan2 k2 &  0.48 (0.043) &  0.26 (0.053) & 0.271 (0.054) \\
 &  &     & scan2 k3 & 0.568 (0.052) & \B 0.454 (0.077) & \B  0.398 (0.04) \\
 &  &     &   scanLR & \B  0.66 (0.058) & \B  0.446 (0.327) & 0.319 (0.229) \\
 \hline
 \multirow{24}{*}{OPT}  &   \multirow{8}{*}{10\%}  &     \multirow{4}{*}{24} & scan2 k1 &  \B 0.144 (0.01) & 0.365 (0.032) & \B 0.254 (0.015) \\
   &  &     & scan2 k2 & 0.128 (0.011) & \B 0.534 (0.048) & 0.418 (0.012) \\
   &  &     & scan2 k3 & 0.127 (0.013) &  \B  0.58 (0.06) & 0.458 (0.017) \\
   &  &     &   scanLR &  \B 0.145 (0.01) & 0.427 (0.046) & 0.294 (0.023) \\
             \cline{3-7} 
   &  &    \multirow{4}{*}{28}  & scan2 k1 & 0.146 (0.014) & 0.386 (0.065) & \B  0.263 (0.026) \\
   &  &     & scan2 k2 & 0.135 (0.008) & 0.577 (0.061) & 0.427 (0.028) \\
   &  &     & scan2 k3 & 0.133 (0.009) & \B  0.64 (0.054) &  0.48 (0.028) \\
   &  &     &   scanLR & \B  0.193 (0.01) &  0.48 (0.129) & \B 0.252 (0.074) \\
             \cline{2-7} 
   &   \multirow{8}{*}{20\%}  &     \multirow{4}{*}{24} & scan2 k1 & 0.265 (0.022) &  0.35 (0.032) & \B 0.264 (0.015) \\
   &  &     & scan2 k2 & 0.247 (0.015) & 0.522 (0.029) & 0.424 (0.016) \\
   &  &     & scan2 k3 &  0.252 (0.02) &  \B 0.598 (0.06) & 0.473 (0.023) \\
   &  &     &   scanLR & \B 0.315 (0.049) & 0.455 (0.106) & 0.303 (0.092) \\
             \cline{3-7} 
   &  &    \multirow{4}{*}{28}  & scan2 k1 & 0.288 (0.014) & 0.394 (0.039) & \B  0.274 (0.026) \\
   &  &     & scan2 k2 & 0.277 (0.021) & 0.591 (0.057) & 0.429 (0.049) \\
   &  &     & scan2 k3 &  0.266 (0.02) & \B 0.654 (0.055) & 0.492 (0.033) \\
   &  &     &   scanLR & \B 0.384 (0.024) & 0.311 (0.164) & \B 0.159 (0.073) \\
             \cline{2-7} 
   &  \multirow{8}{*}{50\%} &      \multirow{4}{*}{24} & scan2 k1 & 0.594 (0.022) & 0.347 (0.035) & 0.292 (0.031) \\
   &  &     & scan2 k2 &  0.639 (0.02) & 0.472 (0.034) &   0.37 (0.03) \\
   &  &     & scan2 k3 & 0.599 (0.012) & \B 0.634 (0.024) &  \B 0.53 (0.021) \\
   &  &     &   scanLR & \B 0.745 (0.045) & 0.208 (0.022) &  0.14 (0.015) \\
             \cline{3-7} 
   &  &      \multirow{4}{*}{28} & scan2 k1 & 0.629 (0.023) & 0.379 (0.041) & 0.302 (0.033) \\
   &  &     & scan2 k2 & 0.651 (0.026) & 0.501 (0.044) & 0.386 (0.044) \\
   &  &     & scan2 k3 & 0.607 (0.013) & \B 0.658 (0.022) & \B 0.542 (0.022) \\
   &  &     &   scanLR & \B 0.717 (0.036) & 0.239 (0.019) & 0.166 (0.008) \\
\bottomrule
\end{tabular}

    \caption{Results for the \realtoxicity\ dataset. Comparison of our weakly supervised scan methods union of left- and right-tailed $p$-values (\scanlr) and top-{$k$} for $k \in \{1, 2, 3\}$ two-tailed $p$-value scans (\scantwo)
    on auditing BERT and OPT. Mean~(std) results from $10$ random test datasets, with best (significant) results in bold. Results for test sets with different percentages of anomalous data (Anom), different layers (Layers), and size of data flagged/classified anomalous relative to the test dataset (Size).}
    \label{tab:additional-toxicity}
\end{table}

We observe that, similar to the other datasets, precision is at its lowest when dealing with 10\% of anomalous data, but it improves as the amount of anomalous data increases. Additionally, we notice that recall shows an upward trend as the subset size of flagged anomalous data grows, as expected. In general, we observe that $\scantwo\ k3$ exhibits the highest recall, which is expected, as we aggregate results over the top-$3$ outcomes enlarging the subset size.

\subsection{Additional Results on \stereoset}\label{apx:additional-stereoset}
In Table~\ref{tab:additional-stereoset} we report additional information on the performance of our two proposed scanning methods (\scanlr, \scantwo) on the Stereset dataset described in Section~\ref{apx:data}. Note that due to the lack of out-of-distribution data, we do not report a baseline classifier (\clfshift).

\begin{table}
    \centering
    \begin{tabular}{llrllll}
\toprule
 LLM & Anom &  Layer &   Method &     Precision &        Recall &          Size \\
\midrule
\multirow{24}{*}{BERT} &  \multirow{8}{*}{10\%} &      \multirow{4}{*}{10} & scan2 k2 &  \B 0.119 (0.016) & 0.405 (0.084) & \BB 0.168 (0.023) \\
 & &    & scan2 k3 & \B0.123 (0.013) & 0.515 (0.066) &  0.21 (0.016) \\
 & &    & scan2 k4 & \B 0.125 (0.015) & \BB 0.558 (0.065) & 0.223 (0.017) \\
 & &    &   scanLR & \BB 0.136 (0.017) &  0.26 (0.034) & 0.192 (0.007) \\
          \cline{3-7} 
 & &     \multirow{4}{*}{12} & scan2 k2 & \BB 0.156 (0.01) &  0.52 (0.073) & 0.167 (0.024) \\
 & &    & scan2 k3 &  0.13 (0.009) & 0.572 (0.076) & 0.221 (0.024) \\
 & &    & scan2 k4 & 0.129 (0.009) & \BB  0.61 (0.059) & 0.237 (0.016) \\
 & &    &   scanLR &  0.142 (0.04) & 0.245 (0.111) & \BB 0.164 (0.045) \\
  \cline{2-7}
 &   \multirow{8}{*}{ 20\%}  &     \multirow{3}{*}{10} & scan2 k2 & \B 0.255 (0.024) & 0.439 (0.098) &  0.17 (0.028) \\
 & &    & scan2 k3 & \B 0.271 (0.018) & 0.581 (0.062) & 0.215 (0.022) \\
 & &    & scan2 k4 & \B 0.261 (0.017) & \BB 0.621 (0.049) & 0.239 (0.017) \\
 & &    &   scanLR & \BB 0.279 (0.038) & 0.285 (0.038) & \BB 0.204 (0.009) \\
  \cline{3-7}
 & &     \multirow{4}{*}{12} & scan2 k2 & \BB  0.286 (0.018) & 0.521 (0.055) & 0.182 (0.017) \\
 & &    & scan2 k3 & 0.262 (0.023) & 0.594 (0.051) & 0.228 (0.026) \\
 & &    & scan2 k4 & 0.253 (0.013) & \BB 0.644 (0.038) &  0.255 (0.02) \\
 & &    &   scanLR & 0.274 (0.032) & 0.255 (0.085) & \BB  0.183 (0.045) \\
  \cline{2-7}
 &  \multirow{8}{*}{ 50\%} &     \multirow{3}{*}{10} & scan2 k2 & \B 0.626 (0.031) &  0.509 (0.041) &  0.203 (0.012) \\
 & &    & scan2 k3 & \B 0.654 (0.018) & 0.676 (0.043) & 0.258 (0.016) \\
 & &    & scan2 k4 & \B 0.638 (0.018) &\BB  0.721 (0.045) & \BB 0.283 (0.02) \\
 & &    &   scanLR & \BB 0.666 (0.043) &  0.321 (0.04) & 0.242 (0.036) \\
  \cline{3-7}
 & &      \multirow{4}{*}{12} & scan2 k2 & 0.627 (0.015) & 0.564 (0.047) &  0.225 (0.02) \\
 & &    & scan2 k3 & 0.605 (0.013) & 0.629 (0.046) &  0.26 (0.021) \\
 & &    & scan2 k4 & 0.599 (0.013) & \BB 0.649 (0.046) &  0.271 (0.02) \\
 & &    &   scanLR & \BB 0.725 (0.041) & 0.577 (0.071) &  \BB   0.4 (0.06) \\
 \hline
\multirow{24}{*}{OPT}  &  \multirow{8}{*}{10 \%} &      \multirow{4}{*}{24}  & scan2 k2 & \B0.099 (0.009) & 0.432 (0.111) & \BB 0.218 (0.046) \\
& &    & scan2 k3 & \B 0.101 (0.009) &   0.495 (0.1) & 0.246 (0.041) \\
& &    & scan2 k4 & \BB  0.103 (0.011) &\BB  0.532 (0.096) & 0.259 (0.037) \\
& &    &   scanLR & \B0.097 (0.011) & 0.375 (0.072) & 0.389 (0.072) \\
  \cline{3-7}
  & &      \multirow{4}{*}{28} & scan2 k2 & \B 0.124 (0.018) & 0.455 (0.082) & \BB  0.182 (0.013) \\
  & &      & scan2 k3 & \B0.118 (0.014) & 0.515 (0.059) & 0.218 (0.009) \\
  & &      & scan2 k4 & \B0.117 (0.012) &  \BB 0.55 (0.063) & 0.235 (0.008) \\
  & &      &   scanLR & \BB 0.131 (0.029) & 0.395 (0.099) & 0.298 (0.018) \\
  \cline{2-7}
& \multirow{8}{*}{20\%} &  \multirow{4}{*}{24} & scan2 k2 & 0.209 (0.017) & 0.474 (0.131) & \BB 0.226 (0.058) \\

& &    & scan2 k3 & 0.208 (0.018) & 0.526 (0.119) &  0.252 (0.05) \\
& &    & scan2 k4 & 0.207 (0.019) &   0.55 (0.12) & 0.265 (0.046) \\
& &    &   scanLR & \BB 0.266 (0.049) & \BB 0.607 (0.203) & 0.446 (0.081) \\
  \cline{3-7}
    &   &       \multirow{4}{*}{28}  & scan2 k2 & \B 0.237 (0.023) & 0.471 (0.094) & \BB 0.197 (0.026) \\
  &   &      & scan2 k3 & \B 0.234 (0.021) & 0.539 (0.084) &  0.229 (0.02) \\
  &   &     & scan2 k4 &  0.231 (0.02) & \BB 0.568 (0.075) & 0.245 (0.017) \\
  &   &     &   scanLR &  \BB 0.282 (0.05) & 0.495 (0.173) & 0.342 (0.062) \\
  \cline{2-7}
&   \multirow{8}{*}{50\%}  &      \multirow{4}{*}{24}  & scan2 k2 & 0.546 (0.016) & 0.639 (0.141) & 0.291 (0.059) \\
& &    & scan2 k3 & 0.544 (0.019) & 0.662 (0.126) & 0.303 (0.052) \\
& &    & scan2 k4 & 0.539 (0.015) & 0.673 (0.117) & 0.311 (0.049) \\
& &    &   scanLR & \BB 0.878 (0.013) &  \BB 0.93 (0.024) & \BB  0.53 (0.018) \\
  \cline{3-7}
& &       \multirow{4}{*}{28}  & scan2 k2 &  0.556 (0.02) & 0.536 (0.104) & 0.241 (0.042) \\
  &   &      & scan2 k3 & 0.554 (0.016) & 0.583 (0.087) & 0.262 (0.037) \\
  &   &      & scan2 k4 & 0.554 (0.017) &  0.606 (0.08) & 0.273 (0.035) \\
  &   &      &   scanLR & \BB 0.886 (0.009) & \BB 0.912 (0.009) & \BB  0.514 (0.01) \\
\bottomrule
\end{tabular}

    \caption{Results for the \stereoset\ dataset. Comparison of our weakly supervised scan methods union of left- and right-tailed $p$-values (\scanlr) top-{$k$} (for $k \in \{2, 3, 4\}$) two-tailed $p$-value scans (\scantwo).
    Mean~(std) results from $10$ random test datasets, with best (significant) results in bold. Results for test sets with different percentages of anomalous data (Anom), different layers (Layers), and size of data flagged/classified anomalous relative to the test dataset (Size).}
    \label{tab:additional-stereoset}
\end{table}

We observe that, similar to the other datasets, precision is at its lowest when dealing with 10\% of anomalous data, but it improves as the amount of anomalous data increases. For BERT, precision appears higher for \stereoset\ than \realtoxicity\ in the previous section. Additionally, we notice that recall shows an upward trend as the subset size of flagged anomalous data grows, as expected. In general, we observe that $\scantwo\ k3$ exhibits the highest recall, which is expected, as we aggregate results over the top-$3$ outcomes enlarging the subset size.

\subsection{Additional Results on Different Testset Compositions}\label{apx:additional-testset}

In Table~\ref{tab:additional-testset} we report additional information on the performance of our two proposed scanning methods (\scanlr, \scantwo) and the baseline classifier (\clfshift) for test datasets that contain larger amounts of anomalous data (50\%, 80\%, and 95\%). 
We compare the results from the main paper obtained using the Higher Criticism Test, which typically favors smaller subsets, with those obtained using the Berk-Jones Test statistic, which tends to favor larger subset sizes.

\begin{table}
    \centering
    \begin{tabular}{lllllll}
\toprule
        Data & Anom &   Method &  F &     Precision &        Recall &          Size \\
\midrule
\multirow{18}{*}{\rotatebox[origin=c]{90}{\realtoxicity}}  &  \multirow{6}{*}{50\%} &\multirow{2}{*}{\scantwo\  k3}  & bj & 0.597 (0.015) & 0.648 (0.024) & 0.543 (0.021) \\
 &  &  & hc & 0.599 (0.012) & 0.634 (0.024) &  0.53 (0.021) \\
     \cline{3-7} 
 &  & \multirow{2}{*}{\scantwo\  k4}  & bj & 0.584 (0.015) & 0.694 (0.029) & 0.594 (0.025) \\
 &  & & hc & 0.587 (0.013) & 0.672 (0.023) & 0.572 (0.018) \\
     \cline{3-7} 
 &  &   \multirow{2}{*}{\scanlr} & hc &   0.74 (0.04) & 0.196 (0.023) & 0.133 (0.015) \\
 &  &   & bj & 0.743 (0.036) & 0.183 (0.022) & 0.123 (0.014) \\
  \cline{2-7} 
 &  \multirow{6}{*}{80\%} &  \multirow{2}{*}{\scantwo\  k3} & bj & 0.857 (0.009) & 0.652 (0.017) & 0.609 (0.018) \\
 &  & & hc &   0.86 (0.01) & 0.639 (0.021) & 0.595 (0.022) \\
     \cline{3-7} 
 &  & \multirow{2}{*}{\scantwo\  k4}  & bj & 0.849 (0.007) & 0.738 (0.055) & 0.697 (0.055) \\
 &  & & hc & 0.854 (0.012) & 0.678 (0.018) & 0.637 (0.022) \\
     \cline{3-7} 
 &  &   \multirow{2}{*}{\scanlr} & hc & 0.918 (0.026) & 0.188 (0.021) & 0.164 (0.018) \\
 &  &    & bj & 0.919 (0.029) & 0.178 (0.018) & 0.156 (0.014) \\
   \cline{2-7} 
 &  \multirow{6}{*}{90\%} &  \multirow{2}{*}{\scantwo\  k3 }  & bj & 0.928 (0.007) & 0.638 (0.026) &  0.62 (0.026) \\
 &  &  & hc & 0.929 (0.004) &  0.64 (0.044) & 0.621 (0.043) \\
     \cline{3-7} 
 &  & \multirow{2}{*}{\scantwo\  k4 } & bj & 0.927 (0.005) & 0.739 (0.064) & 0.718 (0.062) \\
 &  & & hc & 0.927 (0.006) & 0.677 (0.022) & 0.658 (0.022) \\
     \cline{3-7} 
 &  &   \multirow{2}{*}{\scanlr }  & hc & 0.966 (0.014) &  0.184 (0.02) & 0.172 (0.018) \\
 &  &   & bj & 0.965 (0.015) & 0.177 (0.014) & 0.166 (0.013) \\
 \hline
\multirow{18}{*}{\rotatebox[origin=c]{90}{\stereoset}}  &  \multirow{6}{*}{50\%} &  \multirow{2}{*}{\scantwo\  k3} & bj &  0.54 (0.012) &  0.814 (0.11) & 0.377 (0.053) \\
    &  & & hc & 0.544 (0.019) & 0.662 (0.126) & 0.303 (0.052) \\
         \cline{3-7} 
    &  &  \multirow{2}{*}{\scantwo\  k4} & bj & 0.539 (0.012) & 0.817 (0.108) &  0.38 (0.053) \\
    &  &  & hc & 0.539 (0.015) & 0.673 (0.117) & 0.311 (0.049) \\
         \cline{3-7} 
    &  &    \multirow{2}{*}{\scanlr} & hc & 0.877 (0.015) & 0.943 (0.023) & 0.537 (0.016) \\
    &  &    & bj & 0.884 (0.009) & 0.951 (0.012) & 0.538 (0.007) \\
      \cline{2-7} 
    &  \multirow{6}{*}{80\%} & \multirow{2}{*}{\scantwo\  k3}  & bj & 0.839 (0.011) & 0.745 (0.122) &  0.356 (0.06) \\
    &  &  & hc &  0.847 (0.01) & 0.755 (0.078) & 0.357 (0.035) \\
      \cline{3-7} 
    &  & \multirow{2}{*}{\scantwo\  k4}  & bj & 0.836 (0.011) & 0.783 (0.115) & 0.376 (0.056) \\
    &  &  & hc & 0.846 (0.011) & 0.764 (0.073) & 0.362 (0.032) \\
         \cline{3-7} 
    &  &    \multirow{2}{*}{\scanlr}  & hc &  0.97 (0.005) &  0.936 (0.01) & 0.774 (0.009) \\
    &  &    & bj & 0.975 (0.006) &  0.94 (0.013) & 0.774 (0.012) \\
      \cline{2-7} 
    &  \multirow{6}{*}{90\%} &   \multirow{2}{*}{\scantwo\  k3} & bj & 0.923 (0.007) & 0.792 (0.103) &  0.387 (0.05) \\
    &  &  & hc &  0.931 (0.01) & 0.742 (0.093) & 0.359 (0.043) \\
         \cline{3-7} 
    &  & \multirow{2}{*}{\scantwo\  k4} & bj & 0.923 (0.007) & 0.797 (0.098) &  0.39 (0.048) \\
    &  &  & hc & 0.928 (0.012) &  0.75 (0.089) &  0.364 (0.04) \\
         \cline{3-7} 
    &  &    \multirow{2}{*}{\scanlr} & hc & 0.987 (0.004) &  0.935 (0.01) & 0.855 (0.011) \\
    &  &    & bj & 0.988 (0.005) & 0.942 (0.008) & 0.861 (0.009) \\
\bottomrule
\end{tabular}

    \caption{Results for larger percentages of anomalous data (Anom) in the test set and different test satistics: Higher Criciticsm (hc) and Berk Jones (bj). Results for OPT layer $24$ and our weakly supervised scan methods: union of left- and right-tailed $p$-values (\scanlr), and top-{$k$} for $k \in \{2, 3\}$ two-tailed $p$-value scans (\scantwo). Mean~(std) results from $10$ random test datasets, with best (significant) results in bold. Performance metrics (precision and recall), and size of data flagged anomalous relative to the test dataset (Size).}
    \label{tab:additional-testset}
\end{table}

In our main paper, we exclusively present findings based on test datasets that contain limited quantities of anomalous data. 
Our rationale for this approach stems from the presumption that anomalous data is exceedingly scarce. 
However, our results demonstrate across datasets that the greater the availability of anomalous data during testing, the better our method excels. For example, for \realtoxicity\ data with $90$\% for \scantwo\ top-$4$ scanning with BJ scoring function, which reports $0.9727$ with a recall of $0.739$, while we report approximatly $70$\% anomalous samples, while expecting $80$\%. For \stereoset\, \scanlr\ BJ reports for $90$\% anomalous data approximately $86$\% with precision of $0.988$ and recall of $0.942$. 

In this regard, our method differs from a classifier, the performance of which remains unaffected by the quantity of anomalous data within the test dataset as it operates on a per-sentence basis. 
Our method, instead, takes into consideration the entire dataset, identifying patterns that span across multiple sentences and nodes. 
Our results show that this empowers our approach to uncover patterns in BERT activations that the classifier misses. 
Importantly, even in scenarios characterized by a scarcity of anomalous data, should we require a test dataset containing $50$\% or more anomalous data, our method may still be more data efficient than a classifier. 
While a classifier demands not only anomalous samples in its training data but also a substantial training dataset containing a sufficient number of anomalous samples, our method does not require training data at all.

As an illustration, in line with prior work~\cite{azaria2023internal}, we employed a dataset comprising approximately $50$\% anomalous samples to train their classifier \clfshift. 
Specifically, this training set consisted of $1739$ samples for\hallucinations. 
We tested the classifier and our scanning method on a dataset containing only $10$\% anomalous data, totaling $80$ samples assuming a total of $800$ test samples.
In this case, even if we were to consider a test dataset with $90$\% anomalous data, which would consist of $720$ samples, it becomes apparent that our method requires fewer samples than the process of training the classifier and subsequently testing it on a dataset featuring $10$\% anomalous data.

In the context of auditing, where the objective is to ascertain whether a Language Model (LLM) encodes specific patterns in its internal states, we can indeed assume test datasets tailored for auditing purposes, featuring larger amounts of anomalous data. 
This shows that our method enables more data-efficient and precise pattern detection within LLMs than the classifier.

\subsection{Additional Results for Our \scanlr\ Method}\label{apx:additional-scanlr}
In Table~\ref{tab:additional-scanlr} we report additional information on the performance of our proposed scanning method \scanlr, which unions results from scanning over left-tailed $p$-values (\scanl) and results from scanning over right-tailed $p$-values (\scanr). 
We report not only performance metrics but also the intersection of the returned nodes. 
We report results across all three datasets.

\begin{table}
    \centering
    \begin{tabular}{lllllll}
\toprule
D & A & M &     Precision &        Recall &          Size &           INode \\
\midrule
\multirow{9}{*}{\rotatebox[origin=c]{90}{\hallucinations}} & \multirow{3}{*}{10} &   \scanlshort  &  0.205 (0.27) & 0.264 (0.058) & 0.218 (0.077) &               - \\
 &  &  \scanrshort  &  0.462 (0.37) &  0.329 (0.07) & 0.143 (0.104) &               -  \\
  &  & \scanlrshort  & 0.191 (0.095) & 0.462 (0.131) &  0.29 (0.122) & 113.2 (108.115)  \\
   \cline{2-7} 
&  \multirow{3}{*}{20} &  \scanlshort  &     1.0 (0.0) & 0.257 (0.044) & 0.052 (0.009) &               -  \\
 &  &  \scanrshort  & 0.922 (0.233) &  0.27 (0.033) & 0.075 (0.064) &               -  \\
  &  & \scanlrshort  & 0.931 (0.206) & 0.288 (0.069) & 0.078 (0.074) &     10.5 (31.5)  \\
   \cline{2-7} 
& \multirow{3}{*}{50}  &  \scanlshort  &     1.0 (0.0) & 0.255 (0.023) & 0.128 (0.012) &               -  \\
 &  &  \scanrshort  &     1.0 (0.0) & 0.256 (0.023) & 0.128 (0.011) &               -  \\
  &  & \scanlrshort  &     1.0 (0.0) & 0.256 (0.023) & 0.128 (0.011) &       0.0 (0.0)  \\
  \hline
\multirow{9}{*}{\rotatebox[origin=c]{90}{\realtoxicity}}   &  \multirow{3}{*}{10} &  \scanlshort  & 0.161 (0.019) & 0.379 (0.095) & 0.239 (0.064) &               -  \\
   &  &  \scanrshort  & 0.155 (0.019) &  0.39 (0.045) & 0.252 (0.021) &               - \\
      &  & \scanlrshort  & 0.153 (0.017) &  0.48 (0.087) & 0.312 (0.029) &  650.8 (88.062)  \\
       \cline{2-7} 
  &   \multirow{3}{*}{20}  &  \scanlshort  & 0.325 (0.048) &  0.33 (0.096) & 0.211 (0.073) &               -  \\
   &  &  \scanrshort  & 0.321 (0.045) & 0.369 (0.103) & 0.238 (0.078) &               -  \\
      &  & \scanlrshort  & 0.317 (0.041) &  0.416 (0.13) & 0.272 (0.098) & 406.4 (283.095)  \\
       \cline{2-7} 
  &  \multirow{3}{*}{50}  &  \scanlshort  & 0.742 (0.039) & 0.188 (0.023) & 0.126 (0.015) &               -  \\
   &  &  \scanrshort  & 0.742 (0.038) &  0.191 (0.02) & 0.129 (0.014) &               -  \\
      &  & \scanlrshort  &   0.74 (0.04) & 0.196 (0.023) & 0.133 (0.015) &       0.0 (0.0)  \\
      \hline
   \multirow{9}{*}{\rotatebox[origin=c]{90}{\stereoset}}   &  \multirow{3}{*}{10} &  \scanlshort  & 0.101 (0.017) &  0.29 (0.073) & 0.289 (0.064) &               -  \\
      &  &  \scanrshort  & 0.096 (0.018) & 0.268 (0.052) & 0.279 (0.012) &               -  \\
            &  & \scanlrshort  & 0.092 (0.013) & 0.345 (0.068) & 0.374 (0.055) &   289.3 (58.83)  \\
             \cline{2-7} 
     &  \multirow{3}{*}{20}  &  \scanlshort  & 0.242 (0.065) & 0.407 (0.221) & 0.318 (0.088) &               -  \\
      &  &  \scanrshort  & 0.227 (0.054) & 0.299 (0.084) &  0.261 (0.02) &               -  \\
      &  & \scanlrshort  & 0.235 (0.054) & 0.464 (0.205) & 0.381 (0.087) &   302.6 (78.32)  \\
                \cline{2-7} 
     &  \multirow{3}{*}{50}  &  \scanlshort  & 0.908 (0.011) &  0.899 (0.04) &  0.495 (0.02) &               -  \\
      &  &  \scanrshort  & 0.897 (0.011) & 0.914 (0.029) & 0.509 (0.018) &               -  \\
            &  & \scanlrshort  & 0.877 (0.015) & 0.943 (0.023) & 0.537 (0.016) &      0.6 (1.02)  \\
\bottomrule
\end{tabular}

    \input{}
    \caption{Additional results for our proposed \scanlr\ method. Comparison of scanning over individual left-tailed $p$-values (\scanlshort) and right-tailed $p$-values (\scanrshort) and our proposed union of the results (\scanlrshort). Mean~(std) results from $10$ random test datasets, with best (significant) results in bold. Results for different datasets (D), test sets with different percentages of anomalous data (A) in \%, different layers (Layers). We report performance (precision, recall) and size of data flagged as anomalous relative to the test dataset (Size). For \scanlr, we report the intersection of the returned subset of nodes (INode) out of $4092$.}
    \label{tab:additional-scanlr}
\end{table}

First, we note that the precision of individual scans (\scanlshort\ and \scanrshort) tends to be higher than when considering their union (\scanlrshort), which aligns with expectations since the union incorporates false positive samples from both individual scans. However, as anticipated, the recall increases as we include more true positives from the individual scans. Specifically, we observe that LLM activations for anomalous data exhibit a tendency to shift in both directions relative to the expected ``normal'' distribution. Consequently, the union of scans tends to capture more anomalous data increasing recall at a slight cost to precision.

For instance, in the case of \realtoxicity, when the dataset contains $10$\% anomalous data, we observe a slight drop in precision, going from \scanlshort\ $0.161$ and \scanrshort\ $0.155$ to \scanlrshort\ $0.153$. In contrast, we observe a substantial increase in recall, approximately $10$ percentage points, from \scanlshort\ $0.379$ and \scanrshort\ $0.39$ to \scanlrshort\ $0.48$.

Regarding the intersection of the subset of nodes (INode), we notice a reduction in the node intersection as the proportion of anomalous data in the test set increases. Take, for example, the case of \realtoxicity. When the dataset contains 10\% anomalous data, the subset of nodes returned by individual scans (\scanlshort\ and \scanrshort) exhibits a mean intersection of $113.2$ nodes, with a notable standard deviation of $108.115$. However, for the dataset containing 50\% anomalous data, the intersection dwindles to zero nodes, with no variance observed.
This phenomenon can be attributed to the fact that, when the test set contains only a small fraction of anomalous data, the scans perform weaker in identifying patterns. Conversely, when more anomalous data is present, our scanning methods excel at detecting stronger patterns that span across sentences and nodes.
The left- and right-tailed $p$-values concentrate on extreme deviations to the left and right of the anticipated distribution, respectively. Our results suggest that certain nodes tend to shift activations to the right, while others shift to the left, and each of these shifts is detected separately by the corresponding scans.

\subsection{Additional Results for Our \scantwo\ top-k Method}\label{apx:additional-scantwo}
In Table~\ref{tab:additional-scantwo}, we present additional insights into the performance of our proposed scanning method, \scantwo, which utilizes the top-$k$ approach described in the main paper. 
In this table, we not only report performance metrics but also showcase the intersection of returned node and sentence subsets for top-$k$ results, with $k$ ranging from $1$ to $5$. 
We report results across all three datasets.

Importantly, in the main paper, we considered $k$ as a hyperparameter and selected the top-$3$ results due to their superior performance. 
In certain practical scenarios, we may aim to assess methods using unlabeled data known to contain anomalous samples, yet lacking specific labels distinguishing ``normal'' from ``anomalous'' instances. 
In this case, determining the optimal value of $k$ may require a different approach. 
To find a suitable value of $k$ in such scenarios, one approach would involve monitoring the evolution of the anomaly score associated with the identified subset as we increment the value of $k$. 
A high score indicates a higher degree of anomaly detected within the subset. 
Therefore, for smaller values of $k$, we anticipate a higher anomaly score. 
As we progressively increase $k$ and consequently remove more data identified as anomalous, the score is expected to decrease. 
At a certain value of $k$, the score may experience a drop, at which point we could select the $k$ just before this decrease in scores as the best choice.

\begin{table}
    \centering

\begin{tabular}{llllllll}
\toprule
          Data & Anom &   $k$ &     Precision &        Recall &      Size (S) &      Size (N) &         INode \\
\midrule
\multirow{15}{*}{\rotatebox[origin=c]{90}{\hallucinations}} & \multirow{5}{*}{10\%} &  1 & 0.813 (0.293) & 0.267 (0.036) & 0.047 (0.042) & 0.149 (0.134) &         0 (0) \\
  &  &  2 & 0.233 (0.118) &  0.502 (0.08) & 0.247 (0.085) &  0.496 (0.09) & 0.523 (0.151) \\
  &  &  3 & 0.149 (0.031) & 0.685 (0.096) & 0.471 (0.094) & 0.747 (0.042) & 0.653 (0.065) \\
  &  &  4 & 0.142 (0.033) & 0.919 (0.069) & 0.675 (0.134) & 0.828 (0.041) & 0.602 (0.045) \\
  &  &  5 & 0.113 (0.014) & 0.946 (0.034) & 0.854 (0.118) & 0.897 (0.033) &  0.57 (0.055) \\
\cline{2-8} 
  &  \multirow{5}{*}{20\%}  &  1 &     1.0 (0.0) & 0.256 (0.043) & 0.051 (0.009) & 0.073 (0.005) &         0 (0) \\
  &  &  2 & 0.711 (0.295) & 0.543 (0.083) &  0.183 (0.08) & 0.301 (0.202) &  0.44 (0.311) \\
  &  &  3 & 0.412 (0.094) & 0.736 (0.072) & 0.379 (0.099) & 0.613 (0.117) & 0.614 (0.131) \\
  &  &  4 & 0.337 (0.027) & 0.899 (0.049) &  0.537 (0.05) & 0.777 (0.021) & 0.672 (0.057) \\
  &  &  5 &  0.26 (0.024) & 0.928 (0.046) &  0.72 (0.072) &  0.86 (0.019) & 0.624 (0.044) \\
\cline{2-8} 
  &  \multirow{5}{*}{50\%} &  1 &     1.0 (0.0) & 0.254 (0.022) & 0.127 (0.011) & 0.074 (0.003) &         0 (0) \\
  &  &  2 &     1.0 (0.0) & 0.512 (0.031) & 0.256 (0.015) & 0.102 (0.004) & 0.544 (0.383) \\
  &  &  3 & 0.779 (0.041) & 0.754 (0.041) & 0.486 (0.039) & 0.508 (0.037) & 0.646 (0.139) \\
  &  &  4 & 0.683 (0.023) & 0.943 (0.032) & 0.692 (0.043) & 0.771 (0.014) &  0.73 (0.048) \\
  &  &  5 &  0.595 (0.04) &  0.969 (0.03) & 0.818 (0.065) & 0.853 (0.011) & 0.656 (0.037) \\
  \hline
\multirow{15}{*}{\rotatebox[origin=c]{90}{\realtoxicity}}  & \multirow{5}{*}{10\%}  &  1 &  0.144 (0.01) & 0.365 (0.032) & 0.254 (0.015) & 0.655 (0.022) &         0 (0) \\
    &  &  2 & 0.128 (0.011) & 0.534 (0.048) & 0.418 (0.012) & 0.773 (0.018) &  0.695 (0.02) \\
    &  &  3 & 0.127 (0.013) &   0.58 (0.06) & 0.458 (0.017) & 0.822 (0.016) & 0.625 (0.028) \\
    &  &  4 & 0.123 (0.013) & 0.604 (0.064) & 0.489 (0.011) & 0.855 (0.015) & 0.574 (0.026) \\
    &  &  5 & 0.121 (0.011) & 0.616 (0.059) &  0.507 (0.01) &  0.873 (0.02) & 0.543 (0.022) \\
\cline{2-8} 
  &  \multirow{5}{*}{20\%} &  1 & 0.265 (0.022) &  0.35 (0.032) & 0.264 (0.015) &  0.65 (0.025) &         0 (0) \\
    &  &  2 & 0.247 (0.015) & 0.522 (0.029) & 0.424 (0.016) & 0.772 (0.019) & 0.671 (0.012) \\
    &  &  3 &  0.252 (0.02) &  0.598 (0.06) & 0.473 (0.023) & 0.815 (0.019) &  0.599 (0.03) \\
    &  &  4 & 0.251 (0.016) & 0.642 (0.054) & 0.512 (0.024) &  0.853 (0.02) & 0.551 (0.019) \\
    &  &  5 & 0.249 (0.011) & 0.669 (0.051) & 0.537 (0.026) & 0.877 (0.019) & 0.524 (0.027) \\
\cline{2-8} 
  &  \multirow{5}{*}{50\%} &  1 & 0.594 (0.022) & 0.347 (0.035) & 0.292 (0.031) &  0.652 (0.02) &         0 (0) \\
    &  &  2 &  0.639 (0.02) & 0.472 (0.034) &   0.37 (0.03) & 0.703 (0.018) & 0.579 (0.013) \\
    &  &  3 & 0.599 (0.012) & 0.634 (0.024) &  0.53 (0.021) & 0.807 (0.013) & 0.589 (0.034) \\
    &  &  4 & 0.587 (0.013) & 0.672 (0.023) & 0.572 (0.018) & 0.854 (0.012) & 0.545 (0.032) \\
    &  &  5 & 0.581 (0.009) &   0.7 (0.015) & 0.602 (0.018) & 0.879 (0.012) &  0.512 (0.03) \\
  \hline
\multirow{15}{*}{\rotatebox[origin=c]{90}{\stereoset}}  & \multirow{5}{*}{10\%} &  1 & 0.101 (0.019) & 0.288 (0.064) & 0.141 (0.011) & 0.418 (0.025) &         0 (0) \\
       &  &  2 & 0.099 (0.009) & 0.432 (0.111) & 0.218 (0.046) & 0.543 (0.022) &  0.648 (0.04) \\
       &  &  3 & 0.101 (0.009) &   0.495 (0.1) & 0.246 (0.041) & 0.619 (0.034) & 0.598 (0.039) \\
       &  &  4 & 0.103 (0.011) & 0.532 (0.096) & 0.259 (0.037) & 0.669 (0.039) & 0.566 (0.037) \\
       &  &  5 & 0.103 (0.011) &  0.552 (0.09) & 0.268 (0.036) & 0.706 (0.038) & 0.537 (0.046) \\
\cline{2-8} 
  &  \multirow{5}{*}{20\%} &  1 & 0.213 (0.031) & 0.298 (0.058) &  0.139 (0.01) & 0.398 (0.024) &         0 (0) \\
       &  &  2 & 0.209 (0.017) & 0.474 (0.131) & 0.226 (0.058) & 0.525 (0.021) & 0.633 (0.039) \\
       &  &  3 & 0.208 (0.018) & 0.526 (0.119) &  0.252 (0.05) & 0.603 (0.039) & 0.584 (0.044) \\
       &  &  4 & 0.207 (0.019) &   0.55 (0.12) & 0.265 (0.046) &  0.655 (0.04) & 0.554 (0.038) \\
       &  &  5 & 0.205 (0.018) & 0.565 (0.111) & 0.275 (0.042) &   0.7 (0.049) &  0.53 (0.038) \\
\cline{2-8} 
  &  \multirow{5}{*}{50\%} &  1 & 0.525 (0.023) & 0.292 (0.021) &  0.14 (0.011) & 0.411 (0.041) &         0 (0) \\
       &  &  2 & 0.546 (0.016) & 0.639 (0.141) & 0.291 (0.059) & 0.518 (0.029) & 0.583 (0.043) \\
       &  &  3 & 0.544 (0.019) & 0.662 (0.126) & 0.303 (0.052) & 0.574 (0.041) & 0.535 (0.052) \\
       &  &  4 & 0.539 (0.015) & 0.673 (0.117) & 0.311 (0.049) & 0.623 (0.052) & 0.513 (0.054) \\
       &  &  5 & 0.537 (0.013) & 0.689 (0.104) &  0.32 (0.044) &  0.66 (0.059) & 0.492 (0.061) \\
\bottomrule
\end{tabular}

    \caption{Additional results for our proposed \scantwo\ top-$k$ method on auditing OPT layer $24$ on \hallucinations\ (hall), \stereoset\ (stereo) and \realtoxicity\ (toxic). Results for \scantwo\ for different $k$. Mean~(std) results from $10$ random test datasets, with best (significant) results in bold. Results for test sets with different percentages of anomalous data (Anom). {We report the subset size of sentences (S) and the subset of nodes (N), as well as the intersection of the returned subset of nodes (INode) between the $k$-th and $(k-1)$-th run.}}
    \label{tab:additional-scantwo}
\end{table}

In terms of performance, for \hallucinations\, precision tends to decrease as the value of $k$ increases, while recall increases with larger values of $k$. This rise in recall is expected, given that a higher value of $k$ leads to the union of more scans, resulting in a larger subset size. Consequently, the larger the subset size, the higher the likelihood of capturing all anomalous data, thus leading to an increase in recall.
The decrease in precision is also expected. At each increment of $k$, we systematically remove the previously identified subset, which we anticipate to contain a significant portion of anomalous data. For instance, when examining \hallucinations\ with a test dataset containing $10$\% of anomalous data, for $k=1$, we observe a mean precision of $0.813$. This suggests that more than $80$ \% of the retrieved instances are true positives. Consequently, for $k=2$, there is a substantial drop in precision as this initially identified subset is removed from the test set. At this point, the test set no longer contains $10$\% of anomalous data but considerably less, given that we have removed a subset roughly equivalent to $5$\% (Size (S))\footnote{Note, this is the aggregated subset size over top-$k$ runs, and not the size returned by the individual runs.} of the training data and consisted of more than $80$\% anomalous data (Precision). Furthermore, the remaining anomalous data in the dataset might not demonstrate as clear patterns as the removed subset, potentially contributing to a decline in precision. Interestingly, for \realtoxicity\ and \stereoset\, we do not necessarily observe such a strong trend, where sometimes, we see an increase in precision for an increase from $k=1$ to $k=2$ or $k=3$ and a consequent drop in precision for larger $k$.

Regarding the size of the subset of nodes returned (Size (N)), we observe that even at a value of $k$ that strikes a balance between precision and recall, a substantial subset of nodes responsible for flagging the union subset as anomalous is still returned. This finding suggests that there may be a sizable subset of nodes responsible for encoding the investigated bias.
For instance, in the case of \hallucinations\ with a test set containing $20$\% anomalous data, top-$3$ results yield a mean precision of $0.412$ and a mean recall of $0.736$, while returning an average of $0.613$ nodes as a subset.

However, this phenomenon can also be explained by the sequential subset removal process. At each step $k$, we might identify a distinct anomalous pattern, potentially encoded in different nodes. Looking at the intersection of nodes between $k$ and $k-1$ for $k \in {2, 3, 4, 5}$, we notice a tendency where approximately $50$\% of the nodes from step $k-1$ tend to be included in step $k$. However, when considering the increase of node subset size (Size (N)) between $k-1$ and $k$, it appears that each $k$ adds a larger number of additional nodes. 
For example, consider the case of \hallucinations\ with a test set containing $20$\% anomalous data. At the top-$2$ level, we find roughly $30$\% of the nodes, while at the top-$3$ level, we identify just above $60$\% of the nodes.

\subsection{Outlook: Analyzing the Returned Subset of Nodes}\label{apx:additional-outlook}

In this section, we show corresponding preliminary results for the potential expansion of our work as discussed in \S~\ref{sec:outlook-tuning}.
As introduced in \S~\ref{sec:method}, our method returns the subset $S^{\star} = Z_S \times O_S$, where $Z_S$ refers to the subset of test samples (sentences) and $O_S$ to the subset of nodes for the layer analyzed. 
In the main paper, we assess whether the subset of sentences $Z_S$, which yield LLM embeddings identified as anomalous, align with the bias under examination and report precision and recall and relative sample subset size. 
In this context, our focus was on bias detection, specifically during the auditing process.

We seek to expand our research by exploring bias mitigation strategies. 
We believe that an examination of the subset of nodes $O_S$ could provide valuable insights. 
$O_S$ represents the collection of nodes that correspond to the most anomalous subset of sentences—those nodes whose empirical $p$-values of activations deviate from the uniform distribution, thereby flagging them as anomalous.
We hypothesize these nodes are pivotal in identifying anomalous patterns within the data, which could guide the efficient fine-tuning of sub-networks for bias mitigation.

\begin{figure}[htbp]
    \centering
    \begin{subfigure}{0.3\textwidth}
        \centering
        \includegraphics[width=\linewidth]{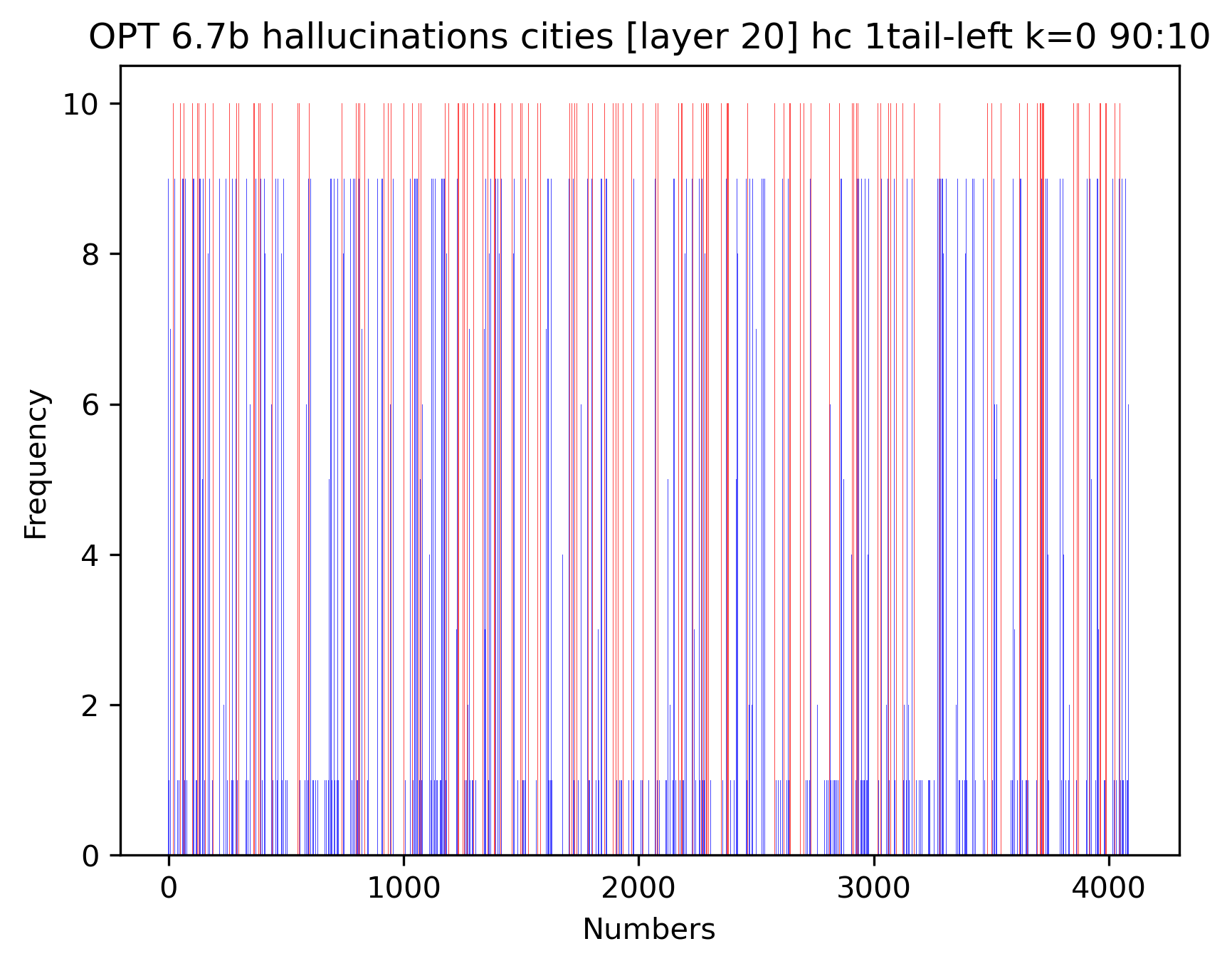}
        \caption{1-tail left, 10\% anom data}
        \label{fig:nodes-1tailleft10}
    \end{subfigure}%
    \begin{subfigure}{0.3\textwidth}
        \centering
        \includegraphics[width=\linewidth]{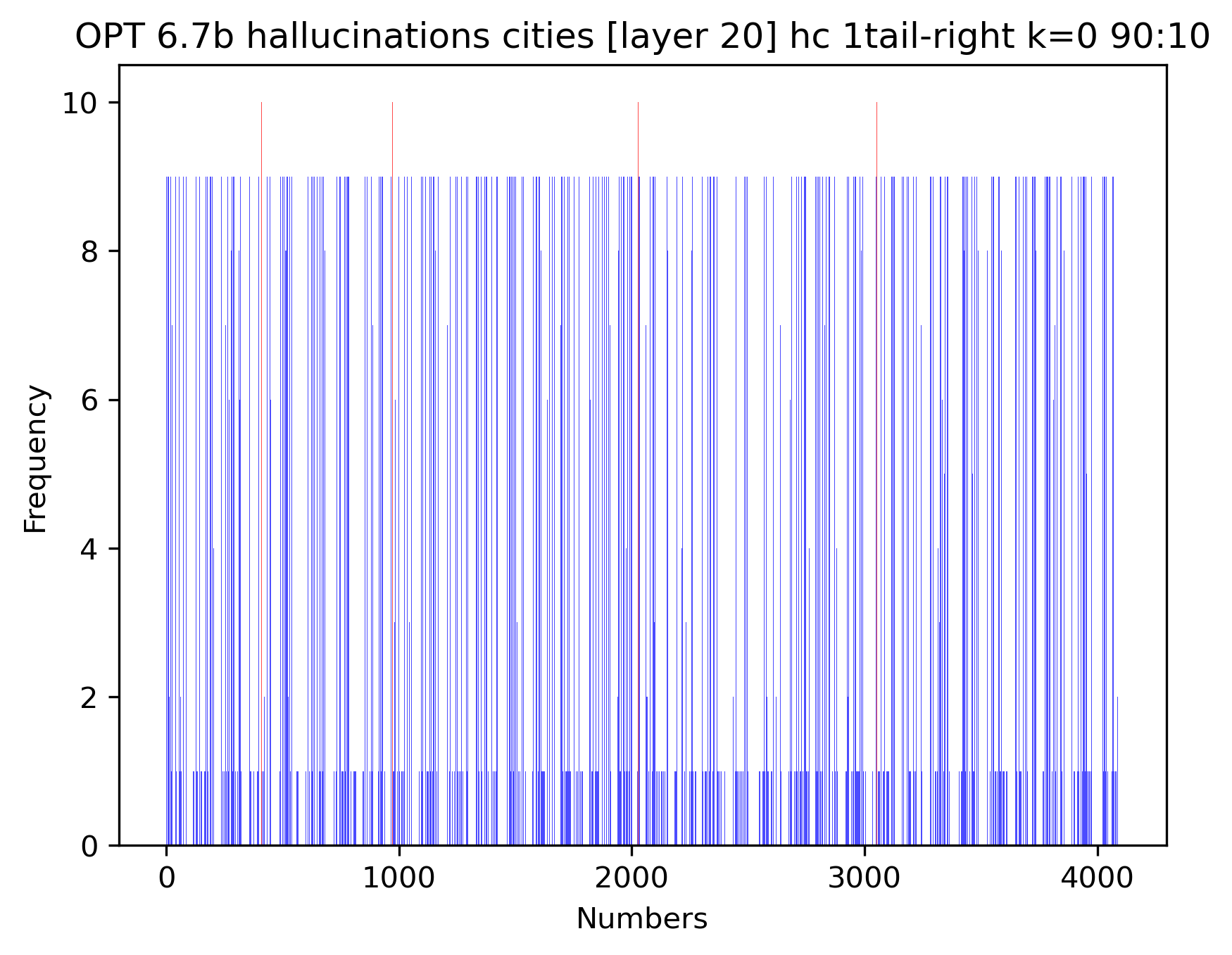}
        \caption{1-tail right, 10\% anom data}
        \label{fig:nodes-1tailright10}
    \end{subfigure}%
    \begin{subfigure}{0.3\textwidth}
        \centering
        \includegraphics[width=\linewidth]{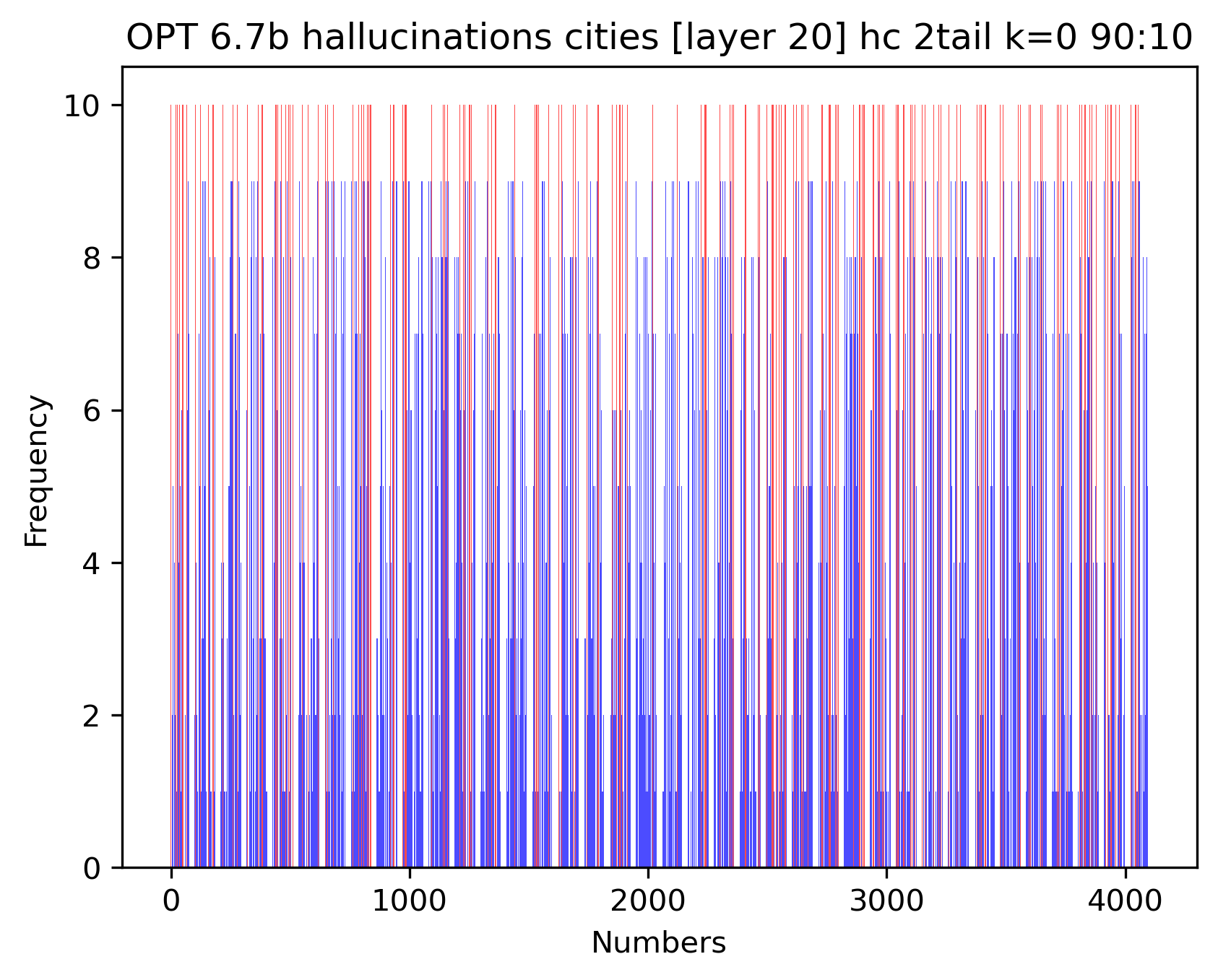}
        \caption{2-tail, 10\% anom data}
        \label{fig:nodes-2tailleft10}
    \end{subfigure}
    \begin{subfigure}{0.3\textwidth}
        \centering
        \includegraphics[width=\linewidth]{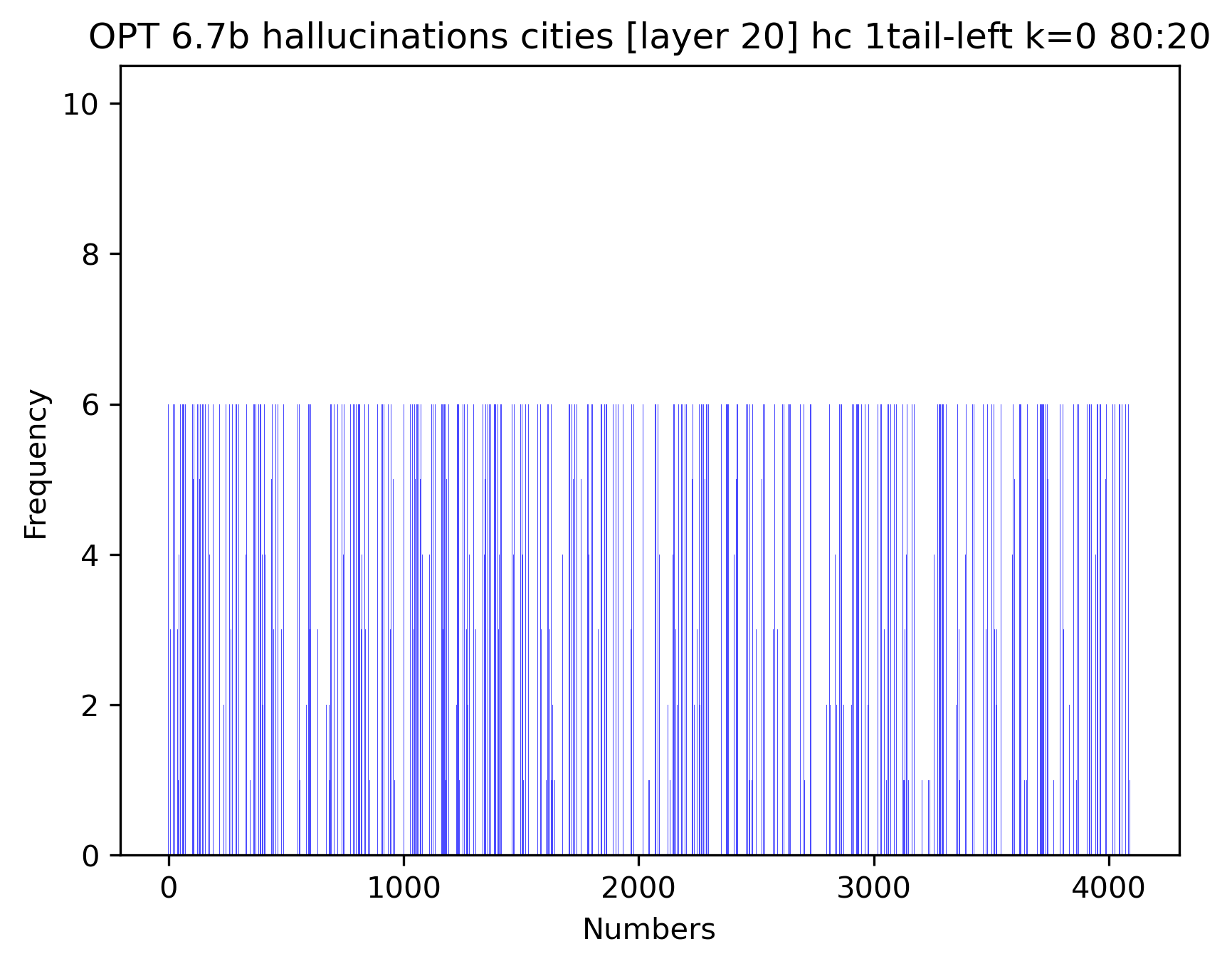}
        \caption{1-tail left, 20\% anom data}
    \end{subfigure}%
    \begin{subfigure}{0.3\textwidth}
        \centering
        \includegraphics[width=\linewidth]{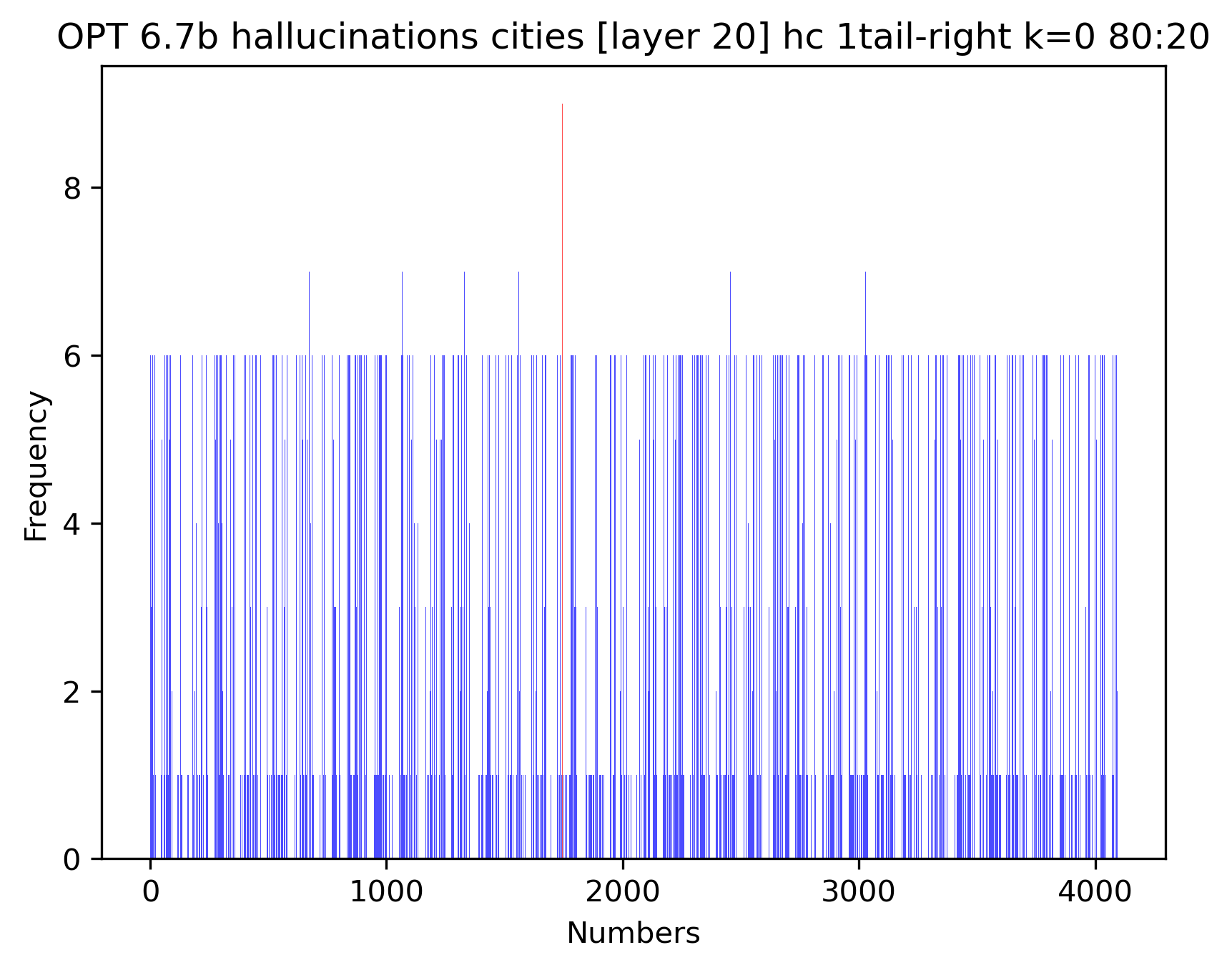}
        \caption{1-tail right, 20\% anom data}
    \end{subfigure}%
    \begin{subfigure}{0.3\textwidth}
        \centering
        \includegraphics[width=\linewidth]{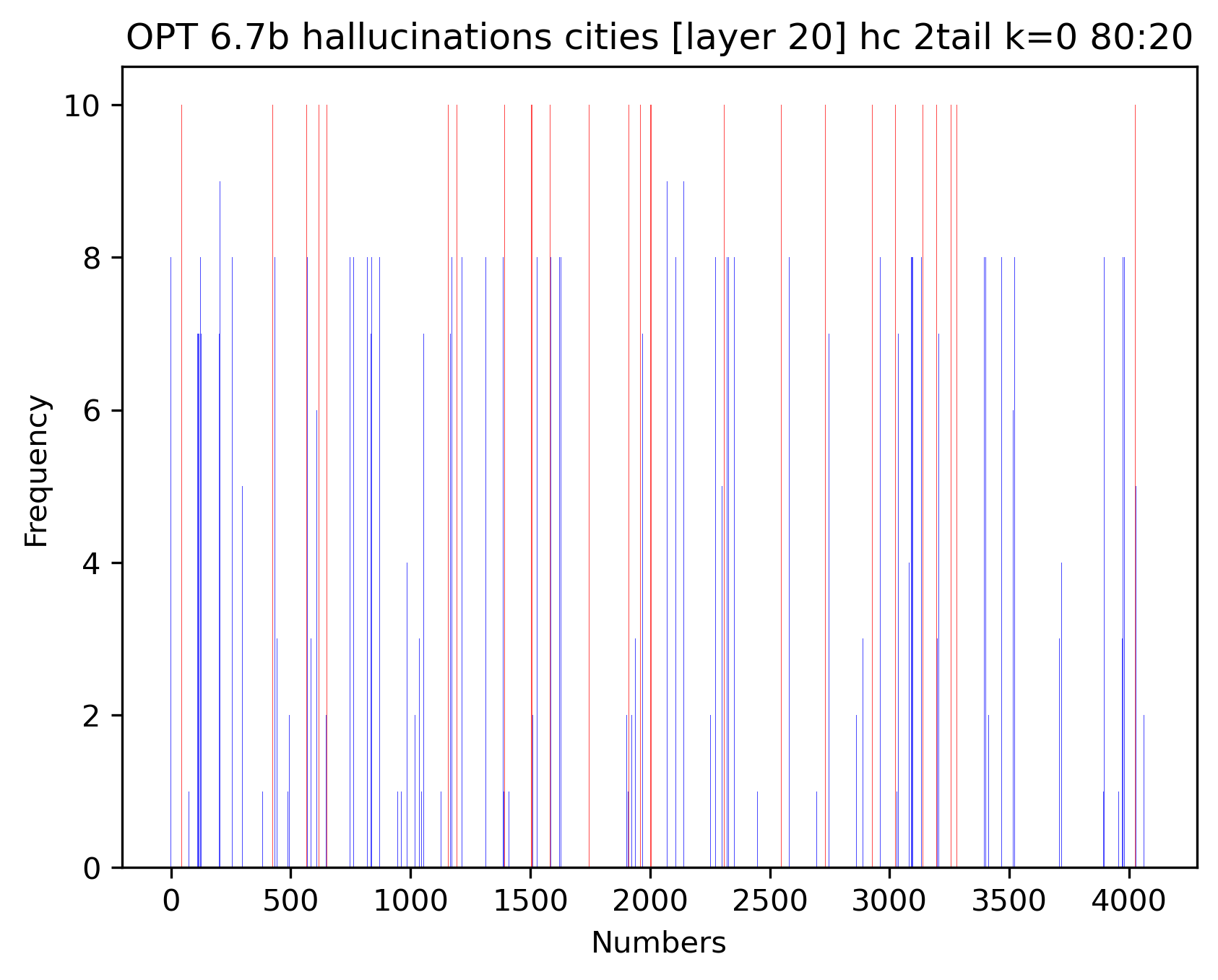}
        \caption{2-tail, 20\% anom data}
    \end{subfigure}
    \begin{subfigure}{0.3\textwidth}
        \centering
        \includegraphics[width=\linewidth]{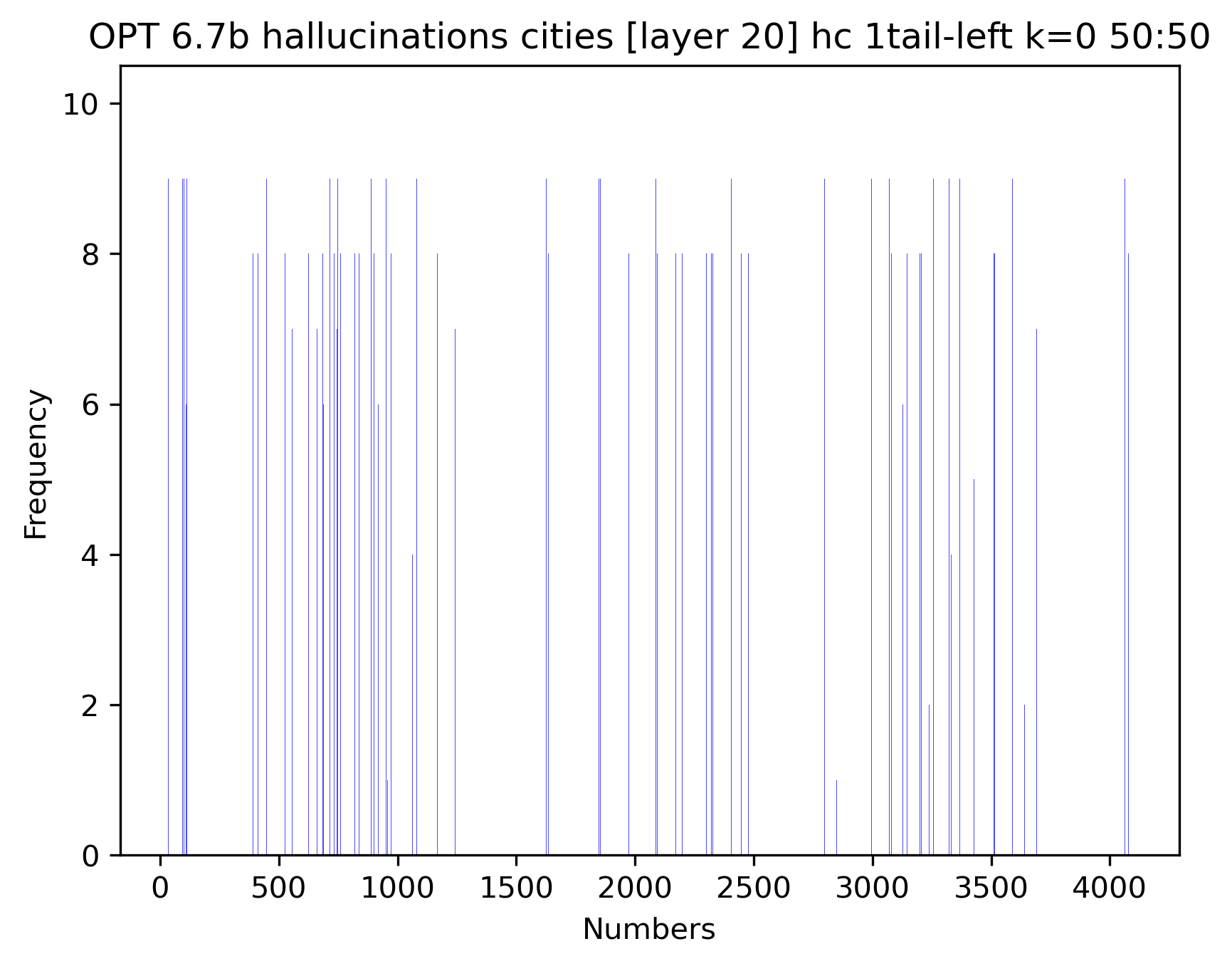}
        \caption{1-tail left, 50\% anom data}
        \label{fig:nodes-1tailleft50}
    \end{subfigure}%
    \begin{subfigure}{0.3\textwidth}
        \centering
        \includegraphics[width=\linewidth]{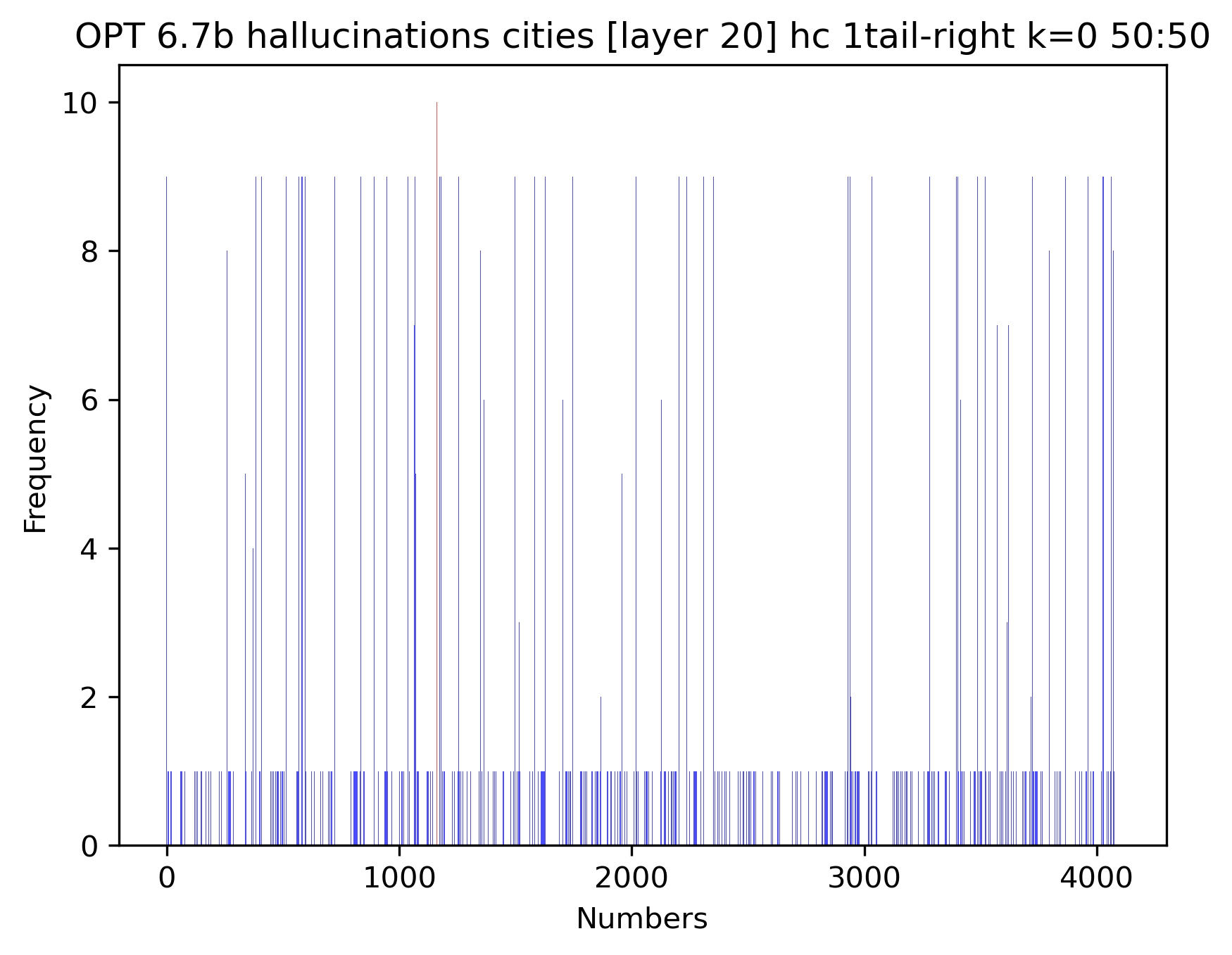}
        \caption{1-tail right, 50\% anom data}
    \end{subfigure}%
    \begin{subfigure}{0.3\textwidth}
        \centering
        \includegraphics[width=\linewidth]{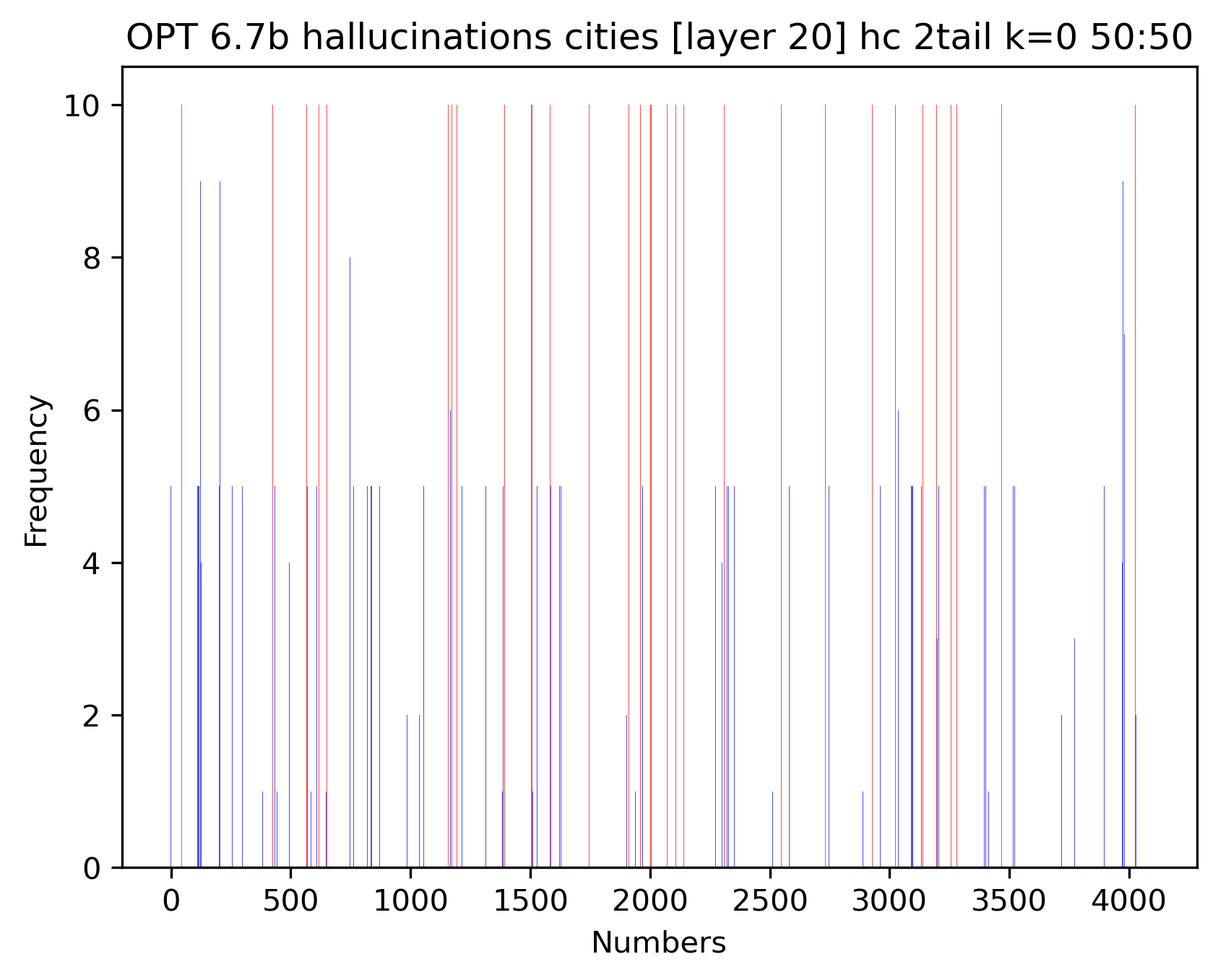}
        \caption{2-tail, 50\% anom data}
    \end{subfigure}
    \begin{subfigure}{0.3\textwidth}
        \centering
        \includegraphics[width=\linewidth]{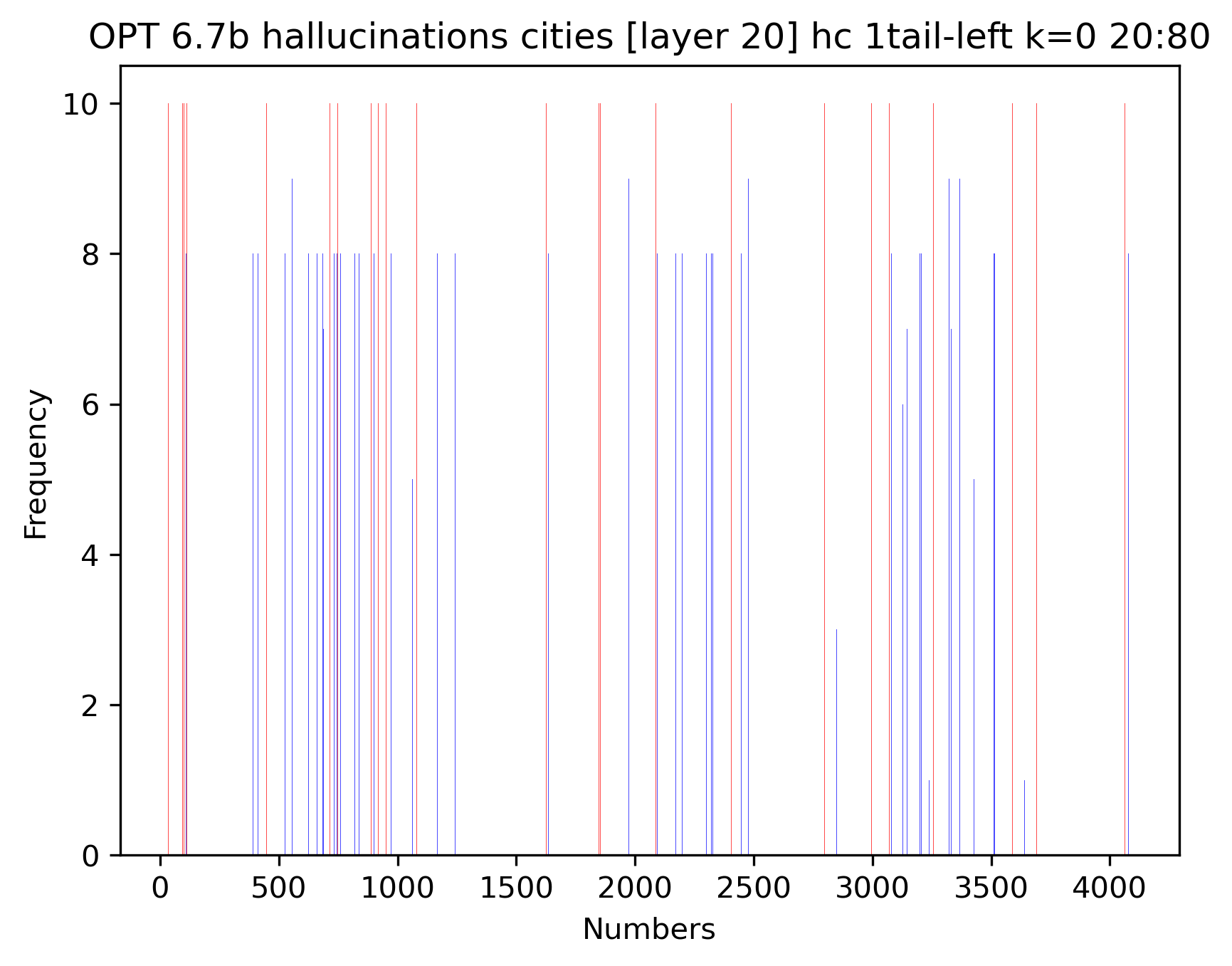}
        \caption{1-tail left, 80\% anom data}
    \end{subfigure}%
    \begin{subfigure}{0.3\textwidth}
        \centering
        \includegraphics[width=\linewidth]{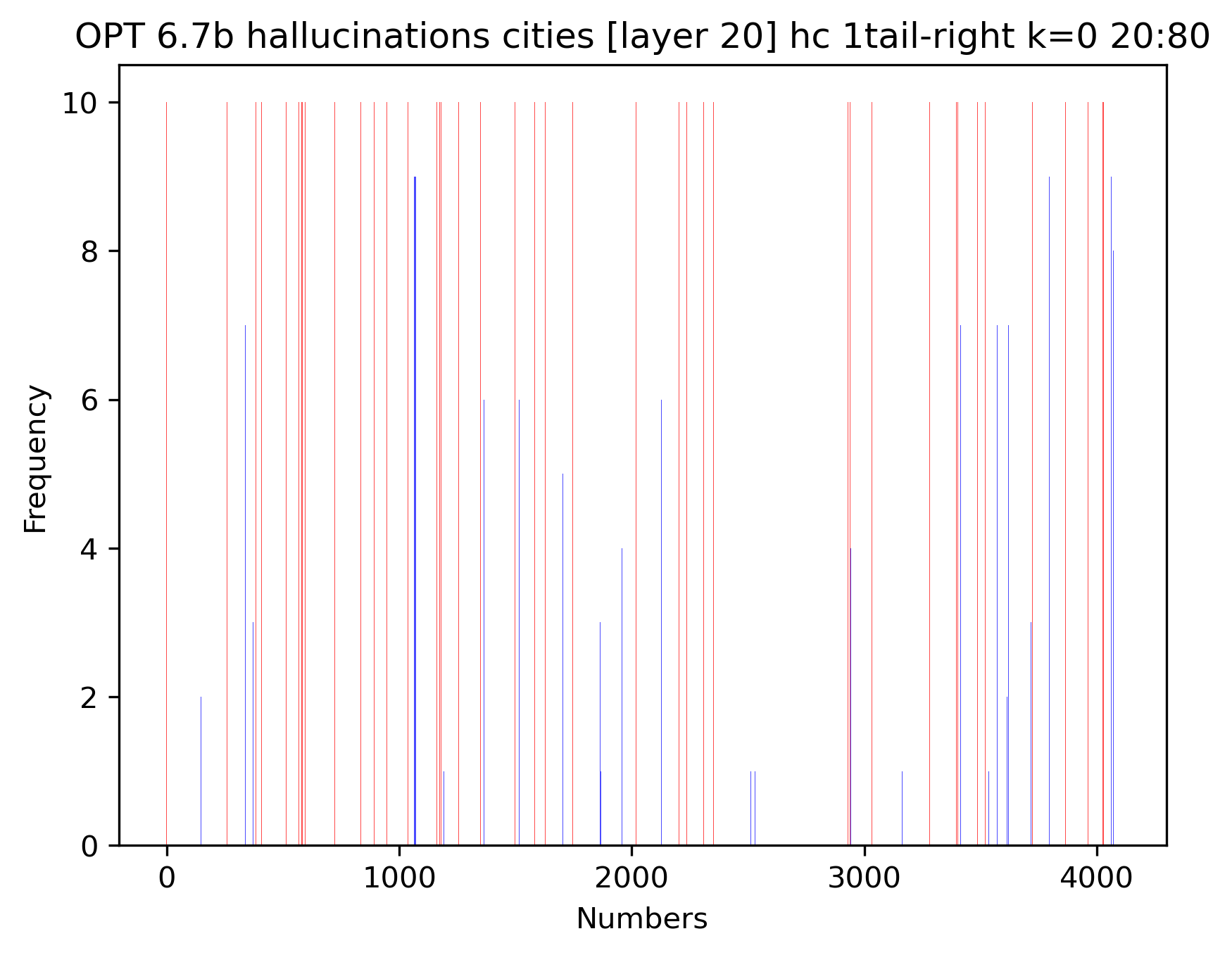}
        \caption{1-tail right, 80\% anom data}
    \end{subfigure}%
    \begin{subfigure}{0.3\textwidth}
        \centering
        \includegraphics[width=\linewidth]{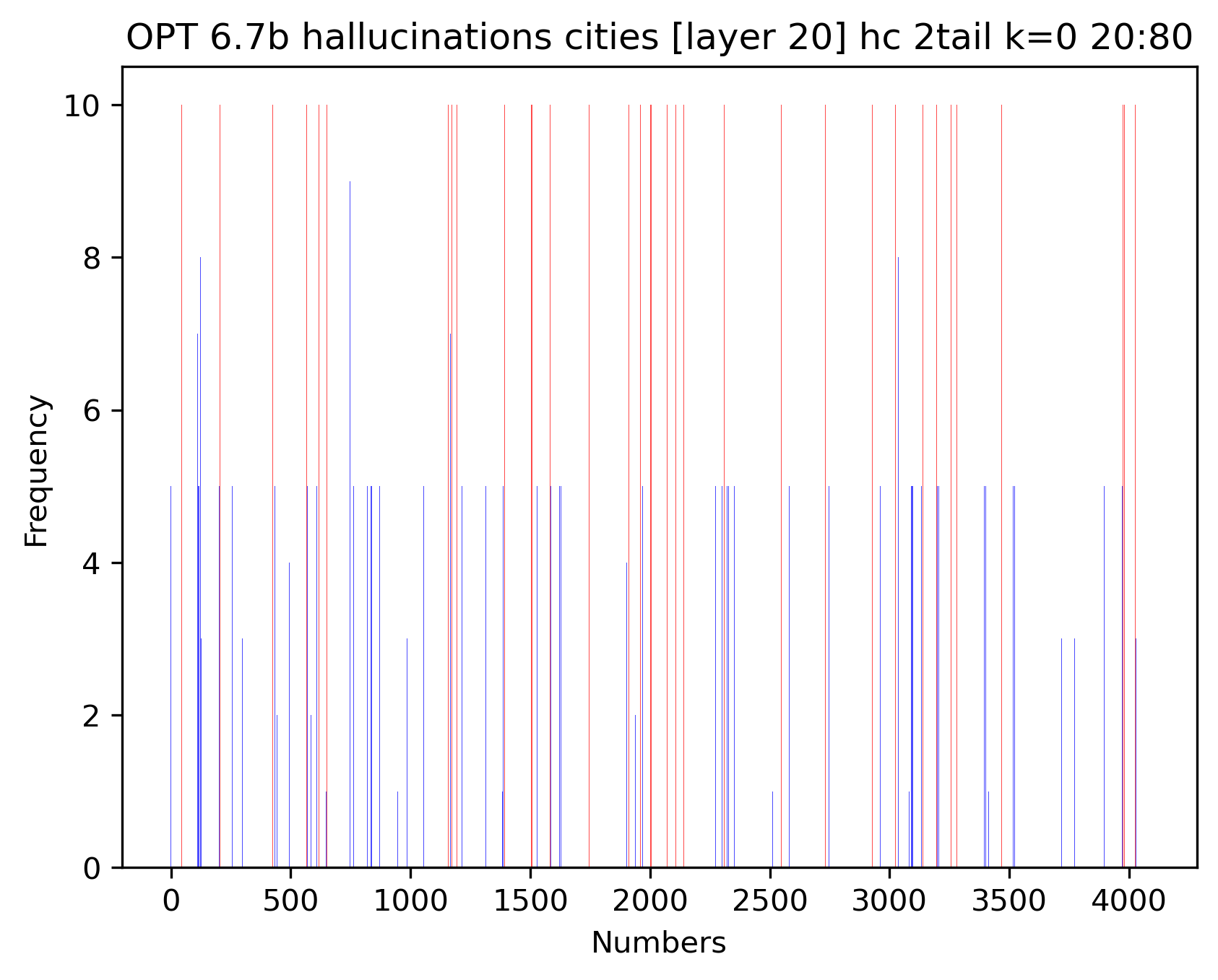}
        \caption{2-tail, 80\% anom data}
    \end{subfigure}
    \begin{subfigure}{0.3\textwidth}
        \centering
        \includegraphics[width=\linewidth]{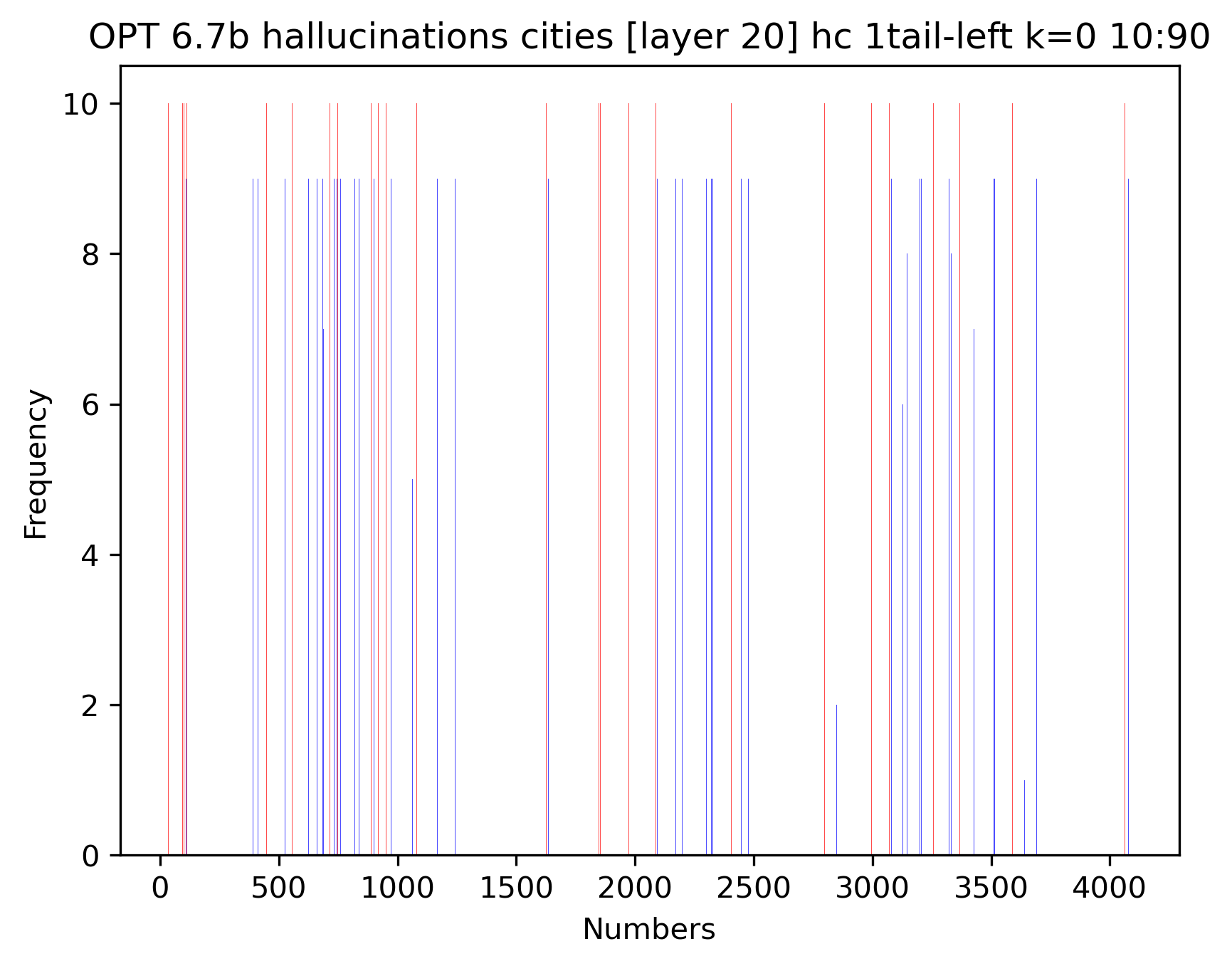}
        \caption{1-tail left, 90\% anom data}
    \end{subfigure}%
    \begin{subfigure}{0.3\textwidth}
        \centering
        \includegraphics[width=\linewidth]{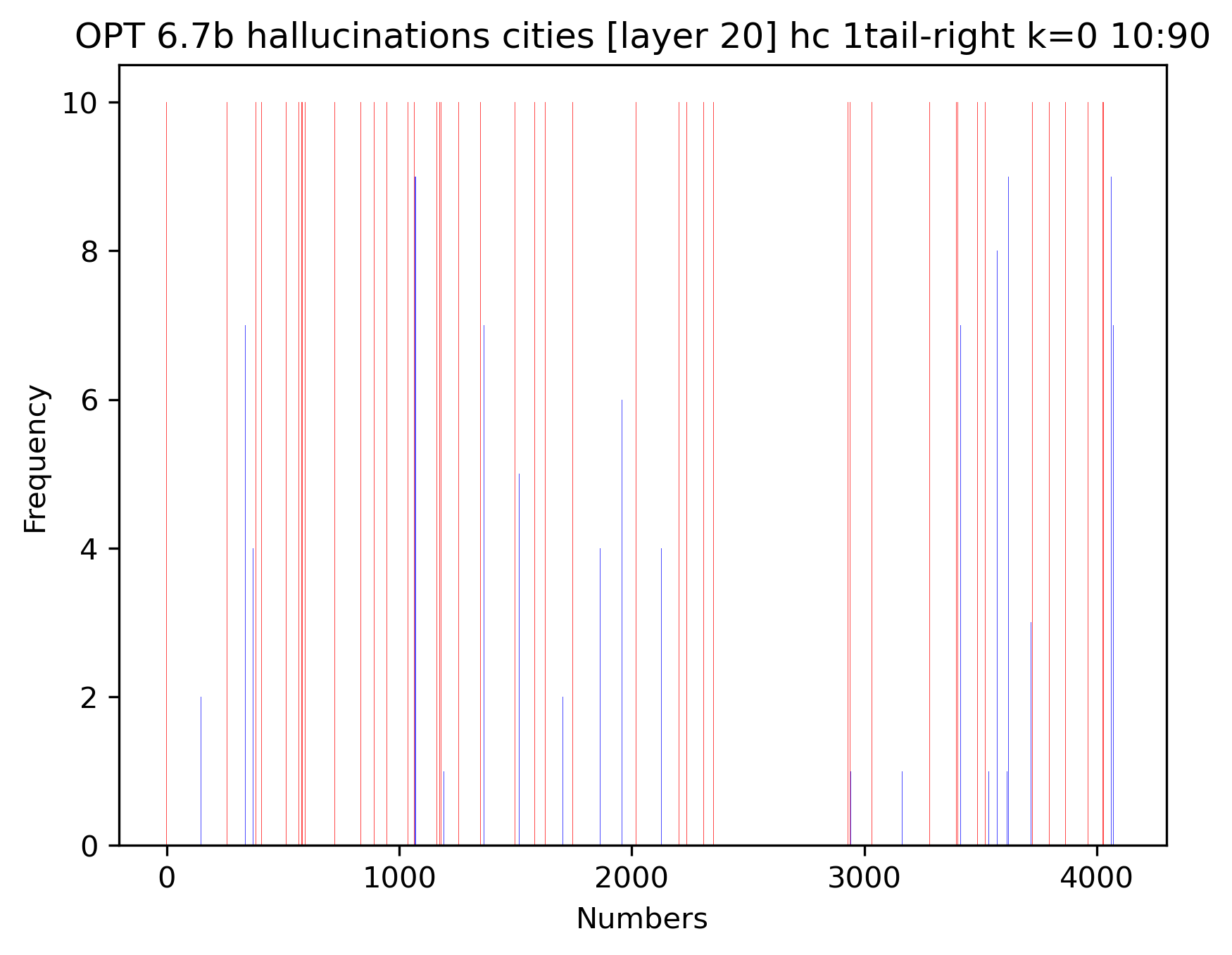}
        \caption{1-tail right, 90\% anom data}
    \end{subfigure}%
    \begin{subfigure}{0.3\textwidth}
        \centering
        \includegraphics[width=\linewidth]{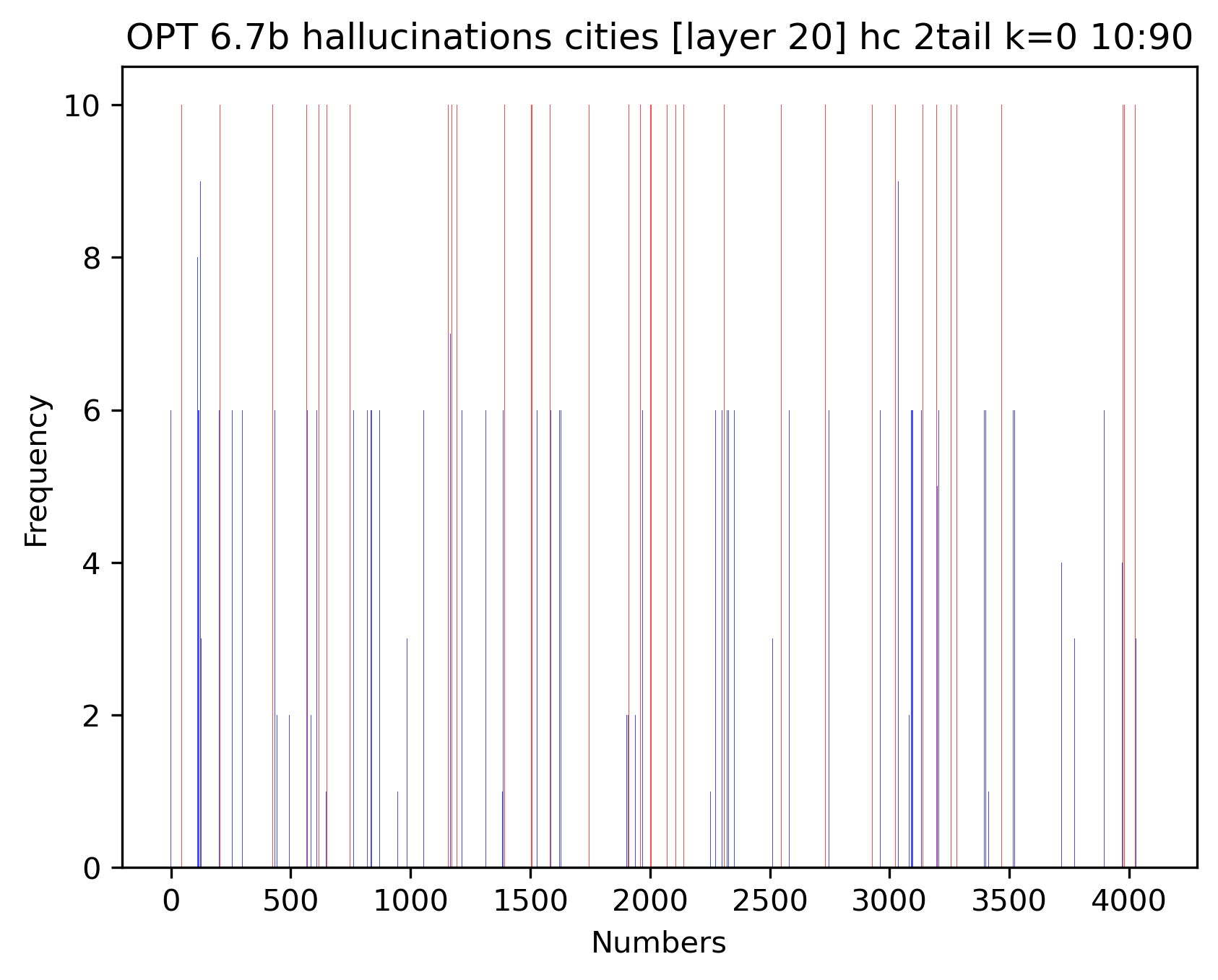}
        \caption{2-tail, 90\% anom data}
          \label{fig:nodes-2tail90}
    \end{subfigure}
    \caption{Frequency of nodes in the returned subset of nodes for test datasets containing different amounts of anomalous data across $10$ random test sets for \hallucinations\ data. Red: nodes that are returned for all $10$ test sets. Results for OPT 6.7b model, activations from layer 20, using the Higher Criticism scoring function. }
    \label{fig:apx_nodes2}
\end{figure}

\paragraph{Exploring the Subset of Nodes} 
In Figure~\ref{fig:apx_nodes2}, we present the node frequency in the subset obtained under varying compositions of test data for the \hallucinations\ dataset. 
The plot illustrates how frequently nodes appear in the resulting subset across $10$ randomly generated test datasets, each containing different percentages of anomalous (false) statements, specifically $10$\%, $20$\%, $50$\%, $80$\%, and $90$\%.
Results are for auditing the OPT 6.7b model, activations from layer $20$, and scanning under the Higher Criticism scoring function for left-tail, right-tail, and top-$1$ 2-tailed $p$-values.

We observe that the lower the amount of anomalous data in the test set, the number of nodes that are returned at high frequency. 
Specifically, when the test data contains only 10\% anomalous data, we consistently find a high number of nodes being returned at least in $9$ out of $10$ random test sets scanned (Fig.~\ref{fig:nodes-1tailleft10} - \ref{fig:nodes-2tailleft10}). 
However, as the percentage of anomalous data in the test set increases to $50$\% or higher (Fig.~\ref{fig:nodes-1tailleft50} - \ref{fig:nodes-2tail90}), we observe a clear reduction in the number of nodes returning at such frequencies. 
Further, in most cases, nodes tend to either appear $9$ or $10$ times or not at all. 
This pattern could be attributed to the scanning method's growing confidence in identifying the anomalous pattern as the subset size increases, leading to a more focused identification of critical nodes.

This observation suggests that when the scanning method gains confidence in pattern detection, it becomes more selective in attributing anomalous activations to a smaller number of nodes. 
Fine-tuning efforts could be efficiently directed toward training these identified critical nodes. 
Note that while this trend holds for various $p$-value calculations, the specific frequencies and numbers of returned nodes vary among the different $p$-value calculations (columns of Fig.~\ref{fig:apx_nodes2}).

\begin{figure}
    \centering
    \begin{subfigure}{0.29\textwidth}
        \centering
        \includegraphics[width=\linewidth]{figures/hallucinations/1-left-20-OPT/nodes_9010.png}
        \caption{Hall, 10\% anom data}
        \label{fig:nodes_hall_10}
    \end{subfigure}%
    \begin{subfigure}{0.28\textwidth}
        \centering
        \includegraphics[width=\linewidth]{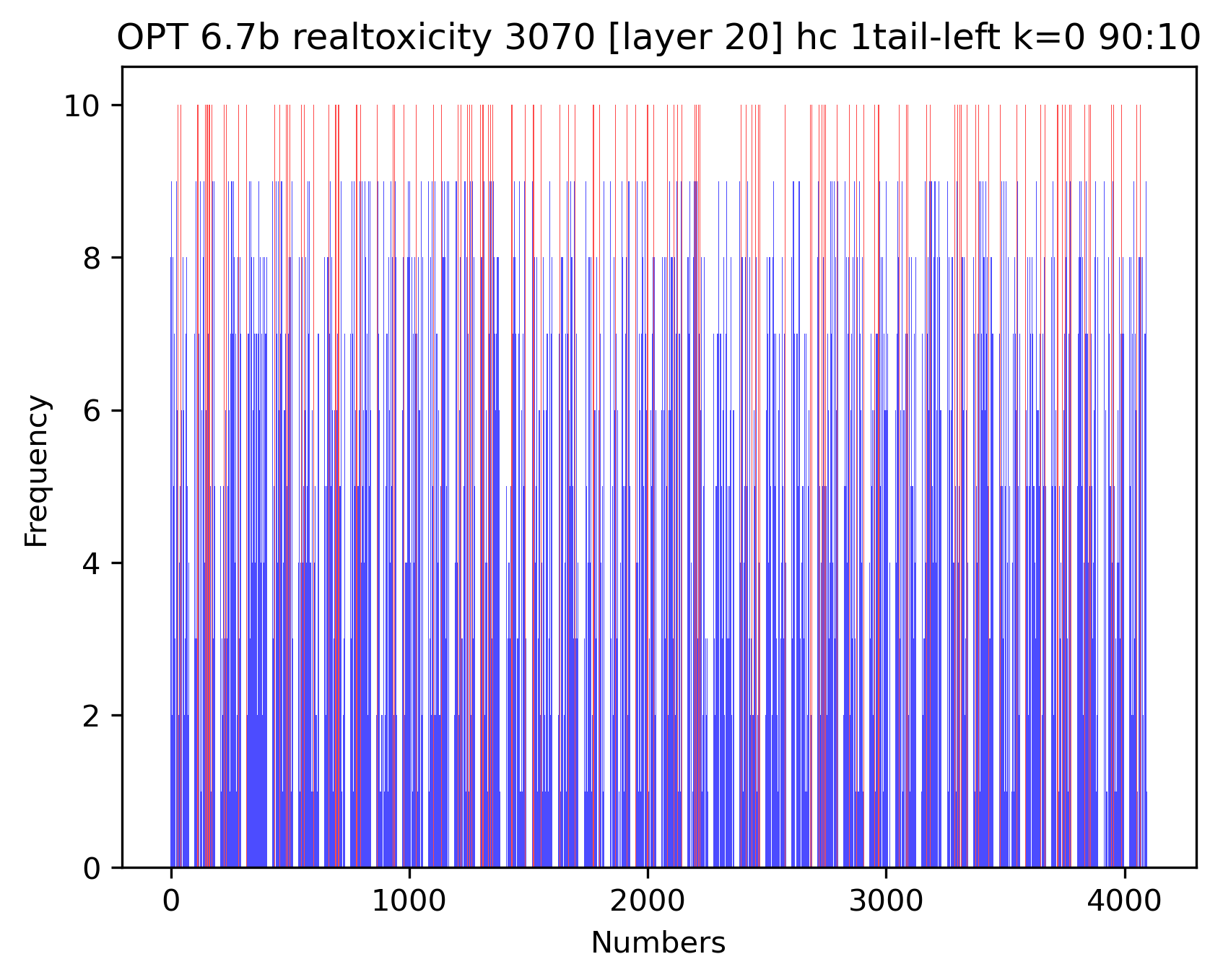}
        \caption{Toxic, 10\% anom data}
        \label{fig:nodes_toxic_10}
    \end{subfigure}%
    \begin{subfigure}{0.3\textwidth}
        \centering
        \includegraphics[width=\linewidth]{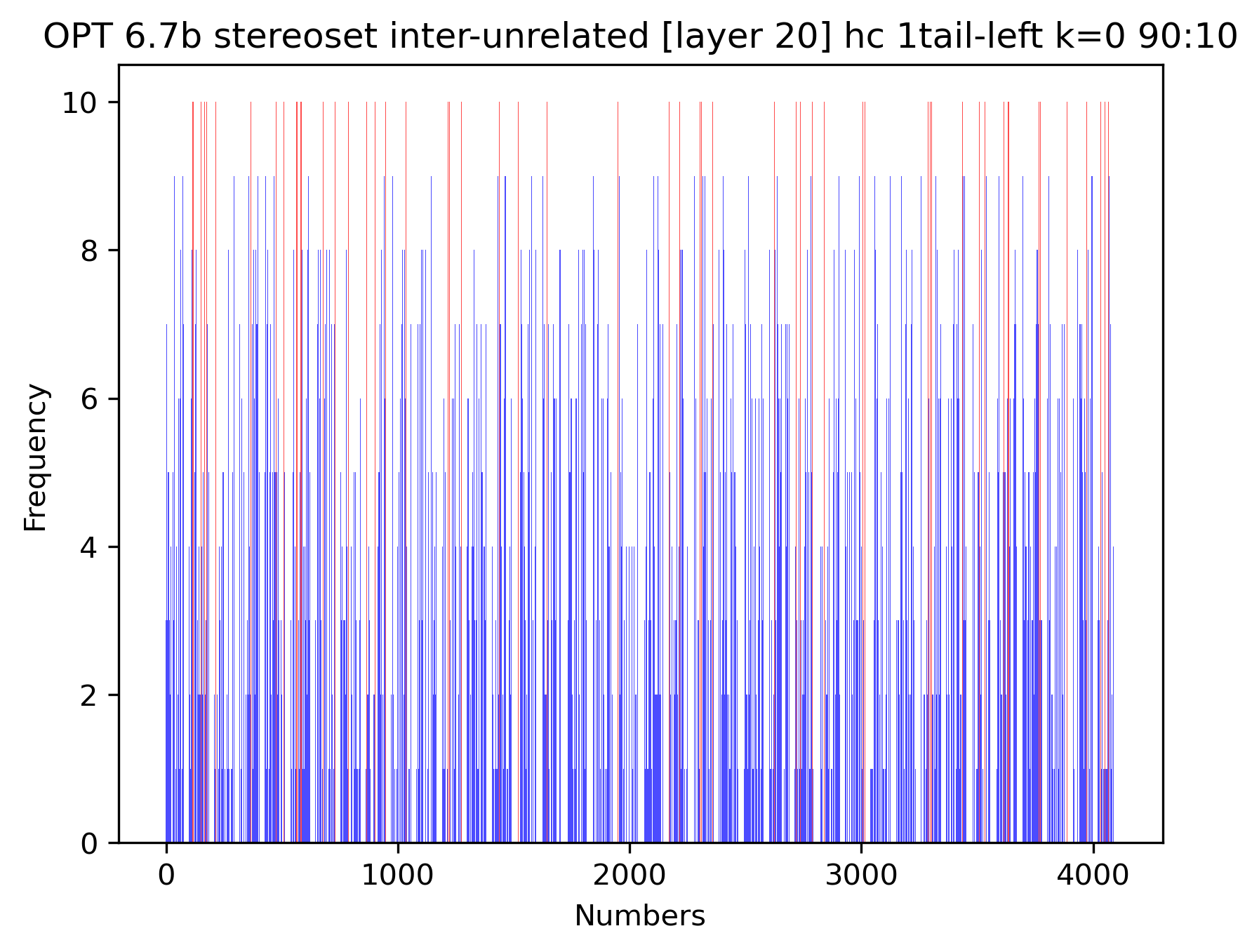}
        \caption{Stereo, 10\% anom data}
        \label{fig:nodes_stereo_10}
    \end{subfigure}
    \begin{subfigure}{0.29\textwidth}
        \centering
        \includegraphics[width=\linewidth]{figures/hallucinations/1-left-20-OPT/nodes_8020.png}
        \caption{Hall, 20\% anom data}
    \end{subfigure}%
    \begin{subfigure}{0.28\textwidth}
        \centering
        \includegraphics[width=\linewidth]{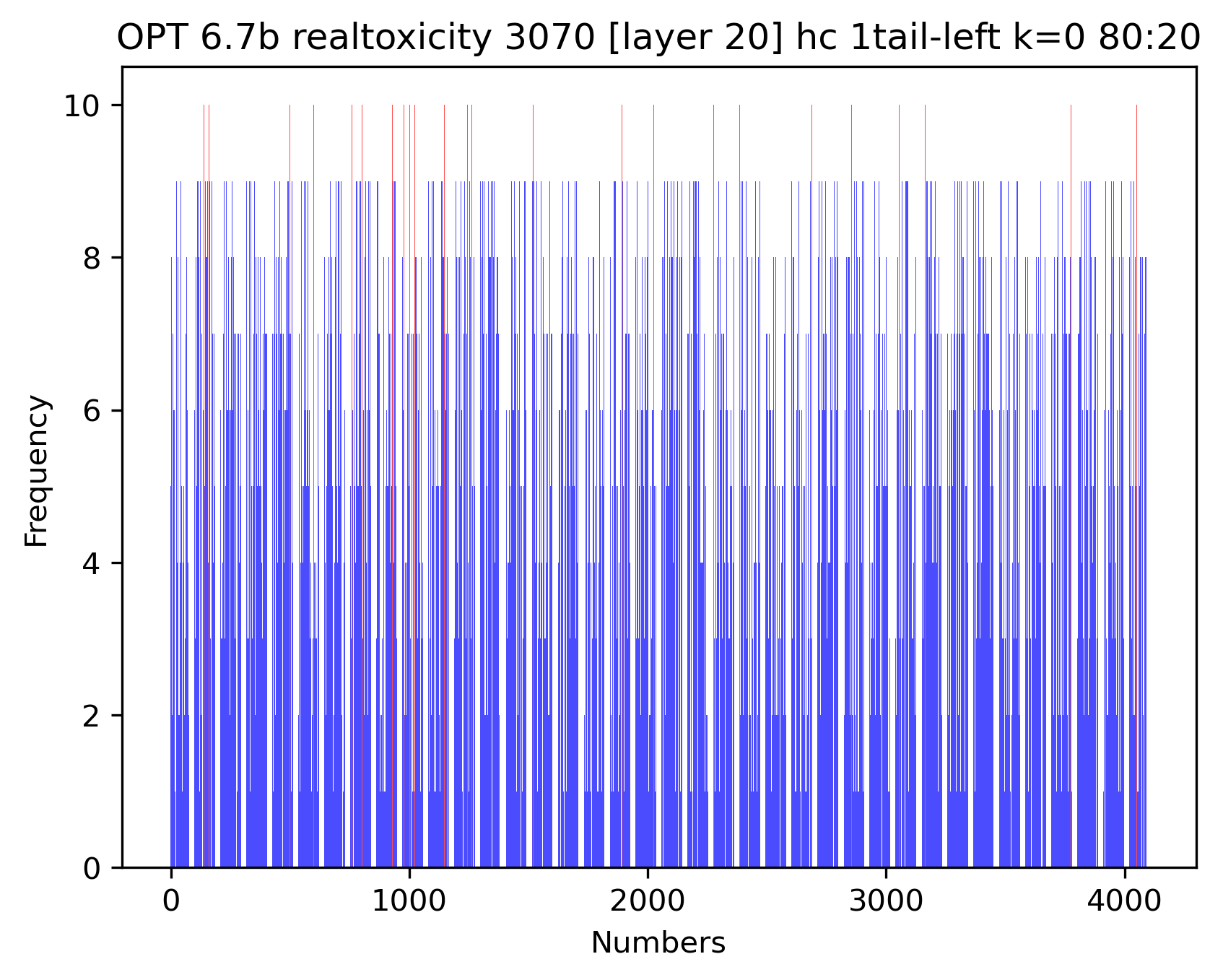}
        \caption{Toxic, 20\% anom data}
    \end{subfigure}%
    \begin{subfigure}{0.3\textwidth}
        \centering
        \includegraphics[width=\linewidth]{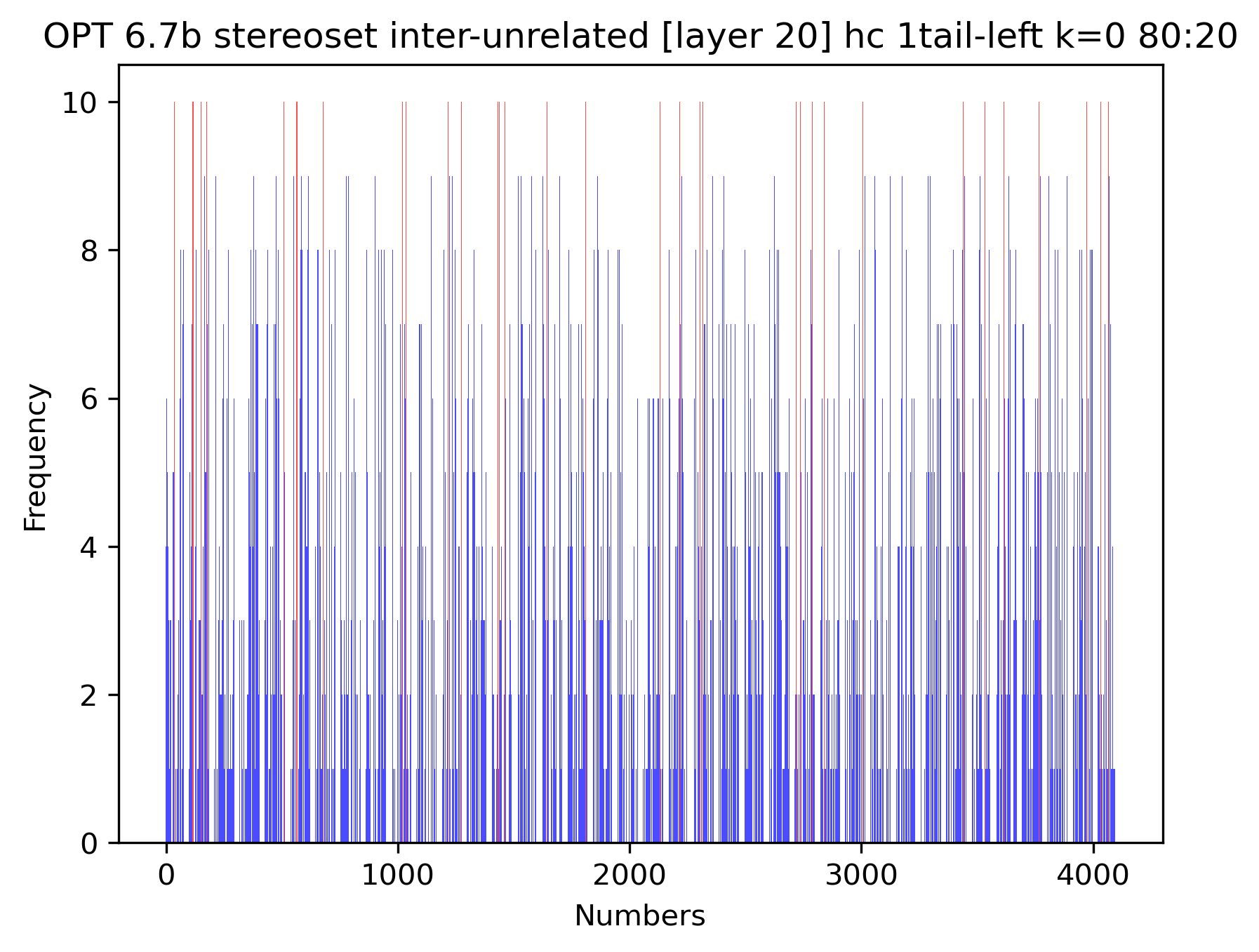}
        \caption{Stereo, 20\% anom data}
    \end{subfigure}
    \begin{subfigure}{0.29\textwidth}
        \centering
        \includegraphics[width=\linewidth]{figures/hallucinations/1-left-20-OPT/nodes_5050.png}
        \caption{Hall, 50\% anom data}
        \label{fig:nodes_hall_50}
    \end{subfigure}%
    \begin{subfigure}{0.28\textwidth}
        \centering
        \includegraphics[width=\linewidth]{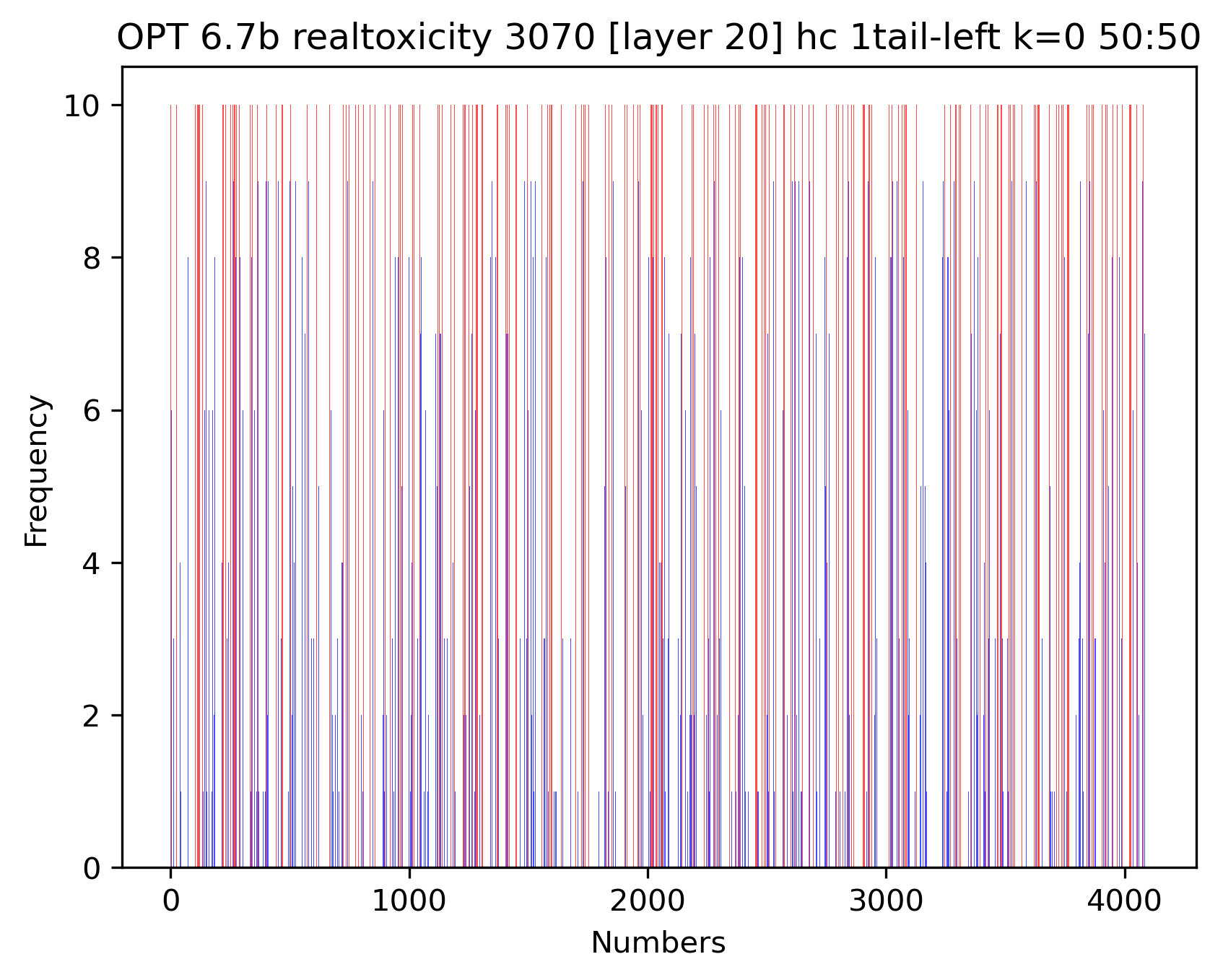}
        \caption{Toxic, 50\% anom data}
    \end{subfigure}%
    \begin{subfigure}{0.3\textwidth}
        \centering
        \includegraphics[width=\linewidth]{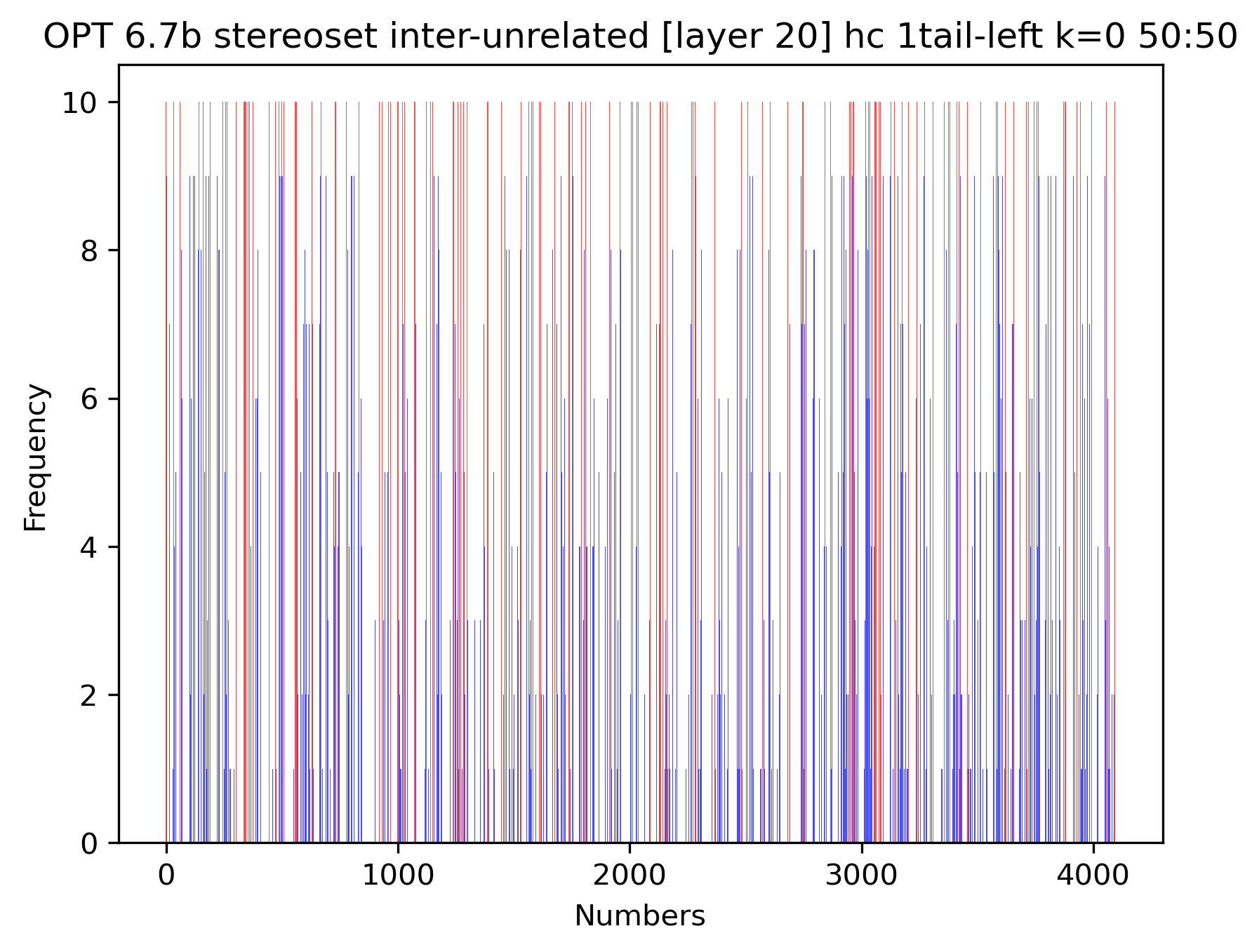}
        \caption{Stereo, 50\% anom data}
    \end{subfigure}
    \begin{subfigure}{0.29\textwidth}
        \centering
        \includegraphics[width=\linewidth]{figures/hallucinations/1-left-20-OPT/nodes_2080.png}
        \caption{Hall, 80\% anom data}
    \end{subfigure}%
    \begin{subfigure}{0.28\textwidth}
        \centering
        \includegraphics[width=\linewidth]{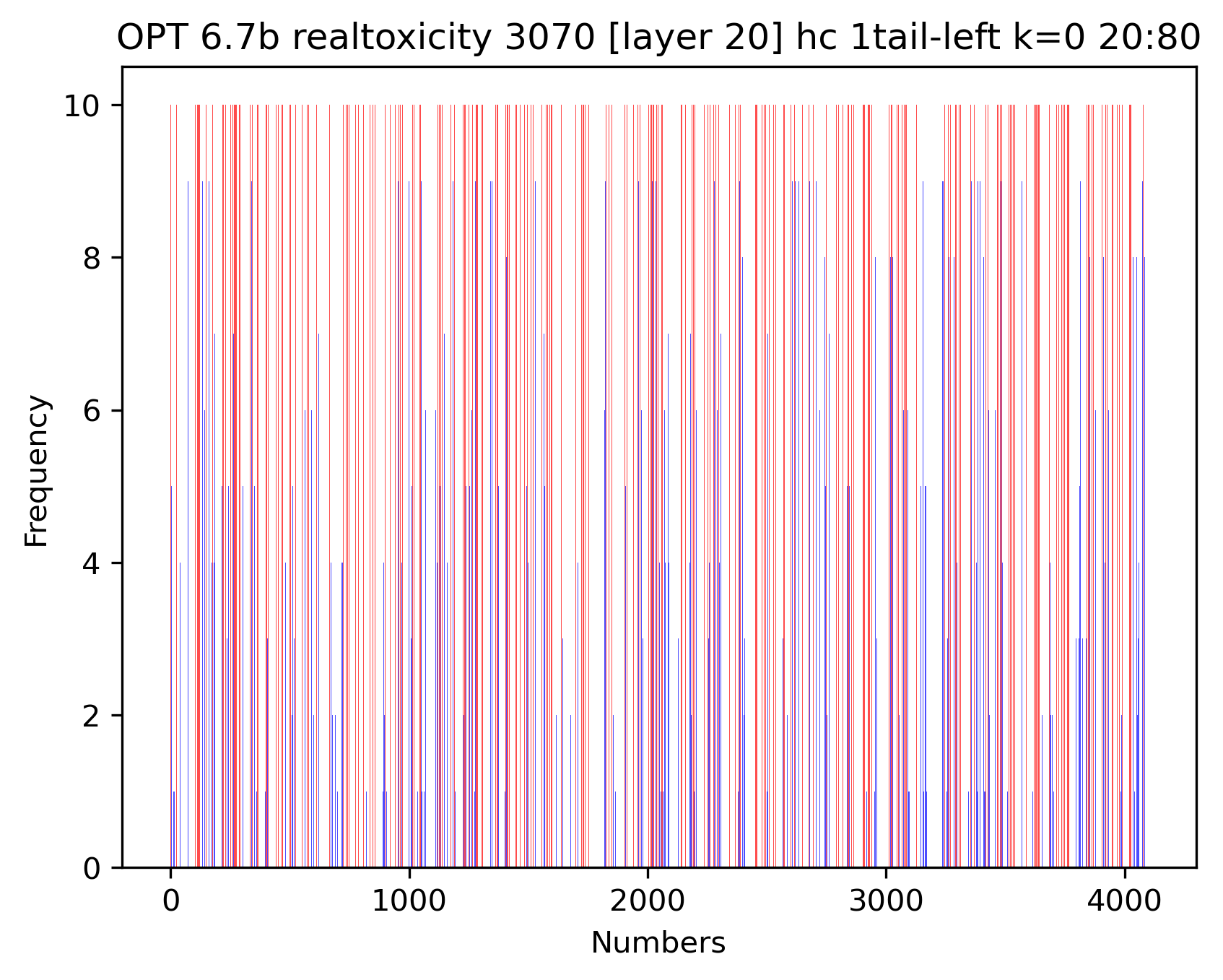}
        \caption{Toxic, 80\% anom data}
    \end{subfigure}%
    \begin{subfigure}{0.3\textwidth}
        \centering
        \includegraphics[width=\linewidth]{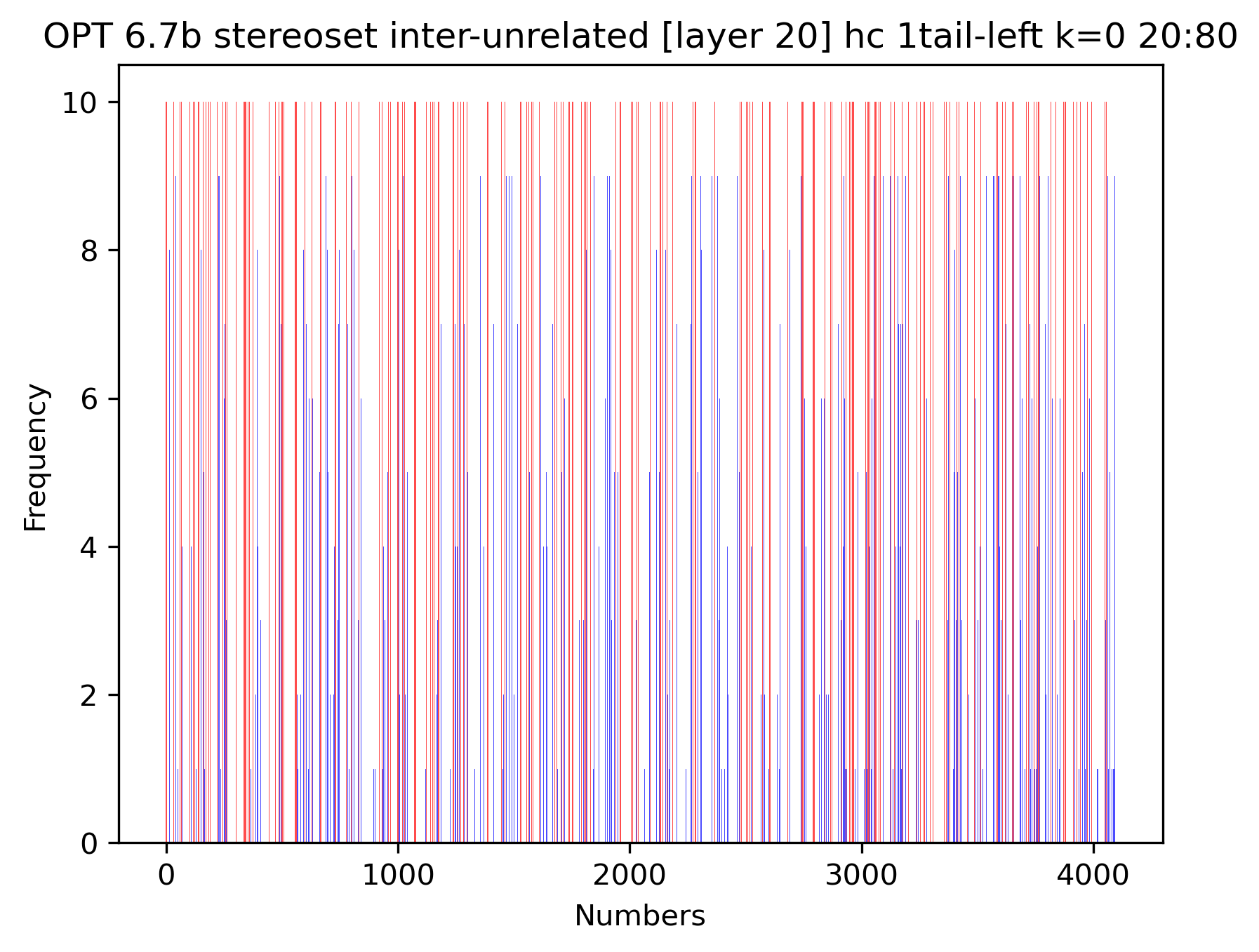}
        \caption{Stereo, 80\% anom data}
    \end{subfigure}
    \begin{subfigure}{0.29\textwidth}
        \centering
        \includegraphics[width=\linewidth]{figures/hallucinations/1-left-20-OPT/nodes_1090.png}
        \caption{Hall, 90\% anom data}
    \end{subfigure}%
    \begin{subfigure}{0.28\textwidth}
        \centering
        \includegraphics[width=\linewidth]{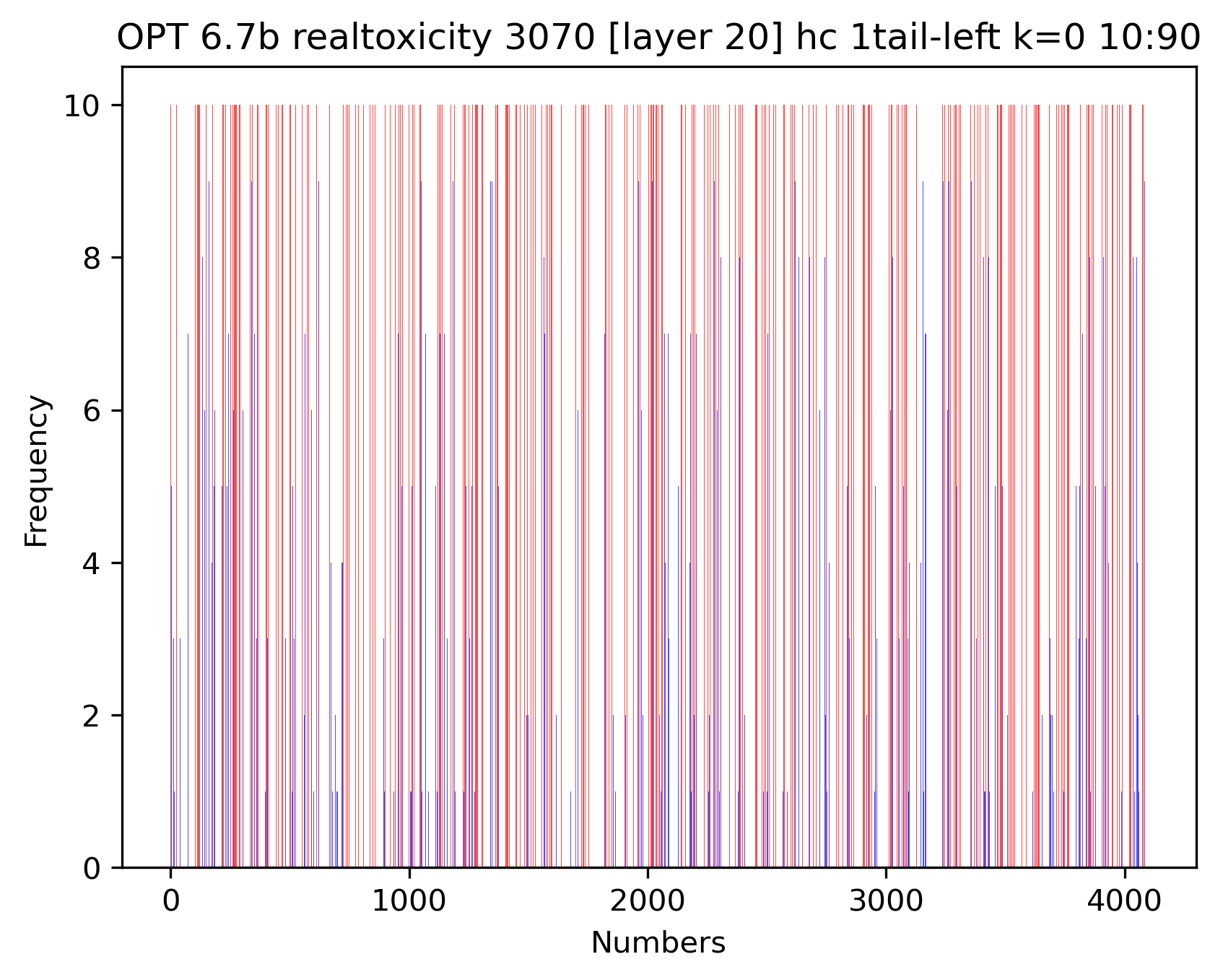}
        \caption{Toxic, 90\% anom data}
    \end{subfigure}%
    \begin{subfigure}{0.3\textwidth}
        \centering
        \includegraphics[width=\linewidth]{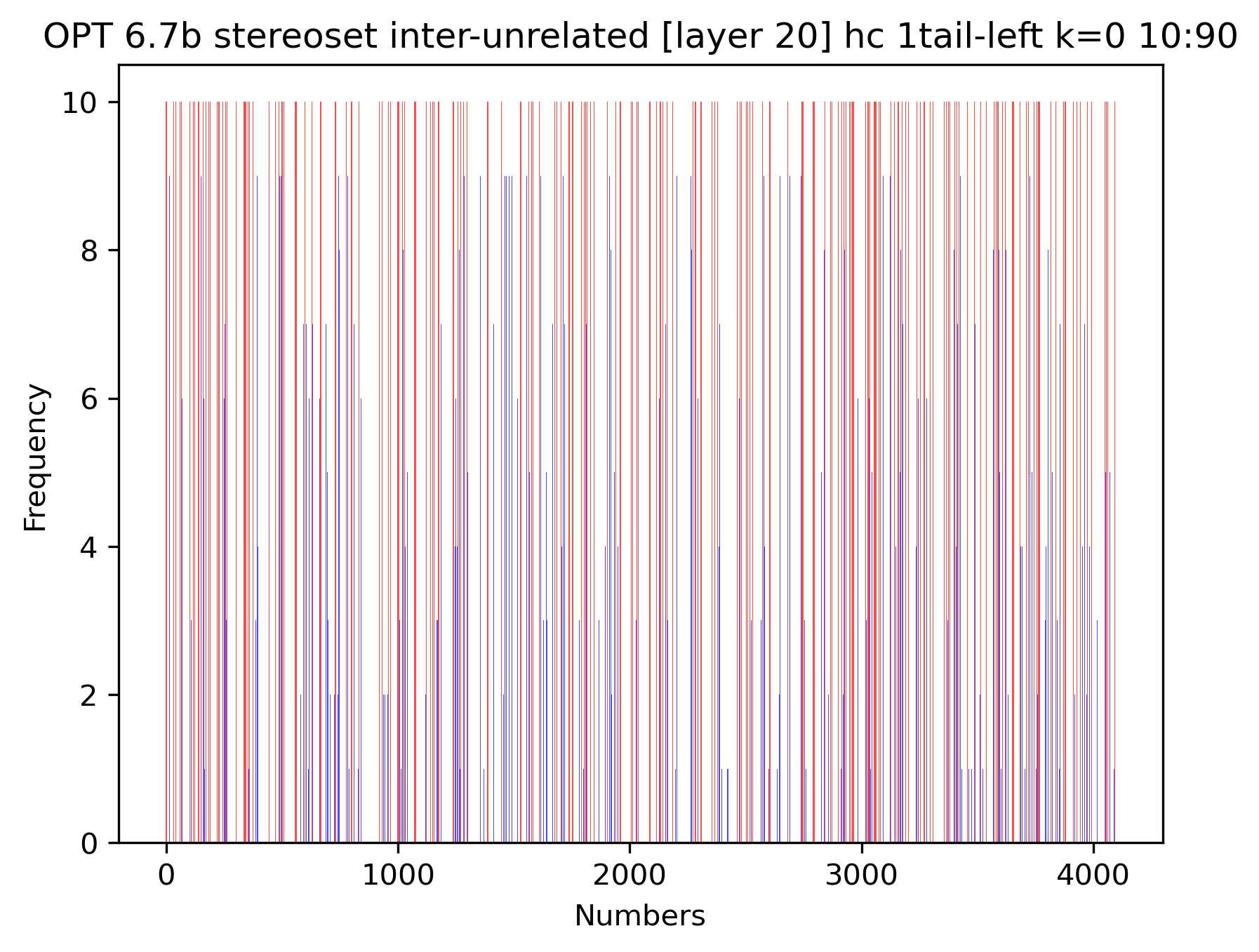}
        \caption{Stereo, 90\% anom data}
          \label{fig:nodes_stereo_90}
    \end{subfigure}
    \caption{Frequency of nodes in the returned subset of nodes for test datasets containing different amounts of anomalous data across $10$ random test sets for \hallucinations\ (Hall), \realtoxicity\ (Toxic), and \stereoset\ (Stereo) data. Results for OPT $6.7$b model, activations from layer $20$, using Higher Criticism scoring function, and left-tailed $p$-values.}
    \label{fig:apx_nodes1}
\end{figure}

We now compare results across datasets.
In Figure~\ref{fig:apx_nodes1} we plot the frequency of each node returned as anomalous for layer $20$ of OPT $6.7$ 2-tailed $p$-values on the \hallucinations, \stereoset, and \realtoxicity\ dataset with a test dataset of $80$\% anomalous data. 
We highlight those nodes that are found in $O_S$ across all $10$ random test sets in red.  

In the case of \realtoxicity\ (toxic) and \stereoset\ (stereo), we observe a consistent trend across datasets: as the quantity of anomalous data increases, more nodes are either included in the anomalous subset across all test sets or are not included at all/included at very low frequencies.

Consistent with our previous observations, this implies that as the scanning method becomes more confident in pattern detection, particularly evident with larger amounts of anomalous data in the test dataset, it tends to exhibit greater selectivity in ascribing anomalous activations. However, notably, in the cases of \realtoxicity\ and \stereoset, this selectivity extends to a larger number of nodes than for \hallucinations.

\paragraph{Informing Fine-Tuning} 
Partial fine-tuning has been proposed as an efficient method that preserves model generalization while customizing pre-trained language models (LLMs) for specific tasks~\cite{tong2023bi,xu2021raise}. 
We believe that identifying pivotal nodes responsible for encoding bias may inform partial fine-tuning processes for bias mitigating in LLMs. 
Note that the definition of bias mitigation depends on the context and stakeholders involved.

For example, to ensure that an LLM remains impartial to stereotypes, we might seek to fine-tune the LLM so that it is indifferent to gender stereotypes, randomly generating both, stereotypical output (e.g., ``She spent too much time on makeup.''~\cite{nadeem2020stereoset}), and anti-stereotypical output (e.g., ``She was fixing her car.''~\cite{nadeem2020stereoset}). 
This process may involve having these pivotal nodes  ``unlearn'' stereotypical patterns.

Conversely, in other (perhaps most) cases, we may aim to empower an LLM to detect instances where it is generating toxic content, hallucinations, or reproducing stereotypes.
In this situation, our focus in the fine-tuning process may be on improving the encoding (or ``learning'') of harmful patterns within these crucial nodes, such that we can subsequently deploy effective bias mitigation strategies. 
This approach would align with the concept that the better we become at identifying bias before releasing generated content or predictions, the more effectively we can provide safer and more responsible model behavior.
\end{document}